%% file: 0.main.tex
\pdfoutput=1

\documentclass[11pt]{article}

\usepackage[]{acl}

\usepackage{times}
\usepackage{latexsym}

\usepackage[T1]{fontenc}

\usepackage[utf8]{inputenc}

\usepackage{microtype}

\usepackage{inconsolata}

\usepackage{amssymb}
\usepackage{graphicx}
\usepackage{booktabs}
\usepackage{multirow}
\usepackage{enumitem}

\title{Revisit Systematic Generalization via Meaningful Learning}

\author{\thanks{$^{*}$ Work was done at Alibaba Group.}$\;\,$Ning Shi$^{\spadesuit\heartsuit}$\qquad \footnotemark[1]$\;\,$Boxin Wang$^{\clubsuit}$\qquad Wei Wang$^{\heartsuit}$\qquad Xiangyu Liu$^{\heartsuit}$\qquad \thanks{$^{*}$ Zhouhan Lin is the corresponding author.}$\;\,$Zhouhan Lin$^{\bigstar}$ \\
  $^{\spadesuit}$Alberta Machine Intelligence Institute, Dept. of Computing Science, University of Alberta \\
  $^{\clubsuit}$University of Illinois at Urbana-Champaign \\
  $^{\heartsuit}$Alibaba Group\qquad $^{\bigstar}$Shanghai Jiao Tong University \\
  \texttt{ning.shi@ualberta.ca, boxinw2@illinois.edu}\\
  \texttt{\{luyang.ww,eason.lxy\}@alibaba-inc.com, lin.zhouhan@gmail.com}
  }

\begin{document}
\maketitle

\input{1.abstract.tex}

\input{2.introduction.tex}

\input{3.meaningful_learning.tex}

\input{4.systematic_generalization.tex}

\input{5.discussion.tex}

\input{6.conclusion.tex}

\input{7.others.tex}

\bibliography{anthology,custom}

\appendix

\input{8.appendix.tex}

\end{document}

%% file: 1.abstract.tex
\begin{abstract}
\vspace{-1mm}
Humans can systematically generalize to novel compositions of existing concepts. Recent studies argue that neural networks appear inherently ineffective in such cognitive capacity, leading to a pessimistic view and a lack of attention to optimistic results. We revisit this controversial topic from the perspective of meaningful learning, an exceptional capability of humans to learn novel concepts by connecting them with known ones. We reassess the compositional skills of sequence-to-sequence models conditioned on the semantic links between new and old concepts. Our observations suggest that models can successfully one-shot generalize to novel concepts and compositions through semantic linking, either inductively or deductively. We demonstrate that prior knowledge plays a key role as well. In addition to synthetic tests, we further conduct proof-of-concept experiments in machine translation and semantic parsing, showing the benefits of meaningful learning in applications. We hope our positive findings will encourage excavating modern neural networks' potential in systematic generalization through more advanced learning schemes.
\vspace{-2mm}
\end{abstract}

%% file: 2.introduction.tex
\vspace{-1mm}
\section{Introduction}
\vspace{-1mm}

\begin{figure}[t]
    \begin{center}
        \includegraphics[width=\linewidth]{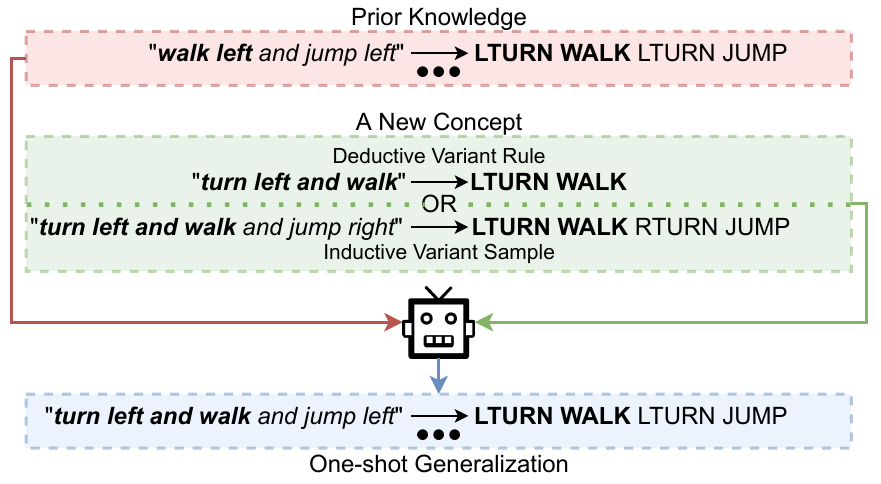}
        \vspace{-8mm}
    \end{center}
    \caption{An example of the one-shot compositional generalization from the old concept ``\textit{\textbf{walk left}}'' to the new one ``\textit{\textbf{turn left and walk}}'' in SCAN. The model is able to generalize from the command ``\textit{\textbf{walk left} and jump left}'' to ``\textit{\textbf{turn left and walk} and jump left}'' through the semantic relationship between the old and new concepts because they refer to the same action ``\textbf{LTURN WALK}''. Such semantic linking can be established by either an inductive sample or a deductive rule.}
    \label{figure:scan}
    \vspace{-5mm}
\end{figure}

As a crucial characteristic of human cognition, systematic generalization reflects people's talents to learn infinite combinations of finite concepts \citep{chomsky1957syntactic,montague1970universal}. Whether connectionist networks can express language and thoughts systematically has been controversial for many years \citep{fodor1988connectionism,hadley1994systematicity,marcus1998rethinking,fodor2002compositionality,brakel2009strong,frank2009connectionist,marcus2018algebraic}. To date, the systematic compositionality in neural networks remains an appealing research topic. Evidence on multiple explicitly proposed language-based generalization challenges suggests that models lack such cognitive capacity \citep{bastings-etal-2018-jump,loula-etal-2018-rearranging,sinha-etal-2019-clutrr,keysers2020measuring,hupkes2020compositionality,kim-linzen-2020-cogs,li-etal-2021-compositional}. Tremendous efforts are made to tackle these challenges through architectural modifications \citep{li-etal-2019-compositional,Gordon2020Permutation,oren-etal-2020-improving,akyurek-andreas-2021-lexicon,chaabouni-etal-2021-transformers}, meta-learning \citep{NEURIPS2019_f4d0e2e7,conklin-etal-2021-meta}, grammar \citep{kim2021sequence,shaw-etal-2021-compositional}, neuro-symbolic models \citep{ChenLYSZ20,NEURIPS2020_83adc922,NEURIPS2020_7a685d9e}, data augmentation \citep{andreas-2020-good,rek2021learning,auersperger-pecina-2021-solving,jiang-bansal-2021-inducing,patel-etal-2022-revisiting}, and loss design \citep{yin-etal-2021-compositional}. Despite their astounding accomplishments, standard sequence-to-sequence (seq2seq) models \citep{sutskever2014sequence} appear to have relatively weak inductive biases, failing to capture underlying hierarchical structure.

\begin{figure*}[t]
    \begin{center}
        \includegraphics[width=\textwidth]{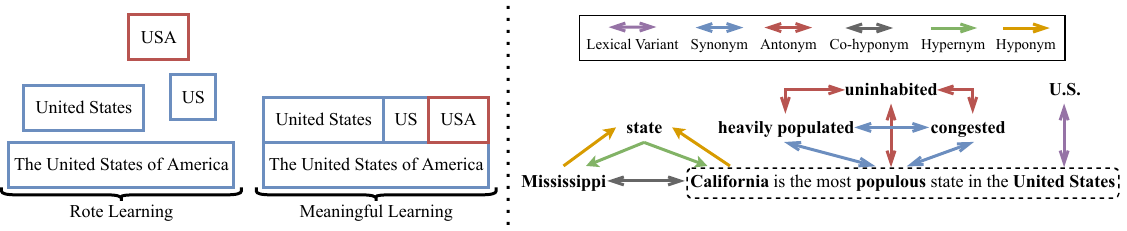}
        \vspace{-8mm}
    \end{center}
    \caption{Adapted examples from Geography. In the left one, intuitively, knowing how the new concept (e.g., ``\textit{USA}'') relates to the other existing ones (e.g., ``\textit{US}'') can boost the learning and memory of this knowledge as a whole. In the right one, bidirectional arrows denote symmetric relations. ``\textit{Mississippi}'' and ``\textit{California}'' are two specific states, and thus both are hyponyms of ``\textit{state}''. In turn, ``\textit{state}'' is a hypernym of them. Due to a common hypernym, ``\textit{Mississippi}'' and ``\textit{California}'' become a co-hyponym for each other. \{``\textit{heavily populated}'', ``\textit{congested}'', ``\textit{populus}''\} is a group of synonyms as sharing similar semantics. Finally, ``\textit{U.S.}'', as a kind of abbreviation, is a lexical variant of ``\textit{United States}''.}
\label{figure:ml}
\vspace{-5mm}
\end{figure*}

In contrast, the successful one-shot generalization in the turn-left experiment on the \textbf{S}implified \textbf{C}omm\textbf{A}I \textbf{N}avigation (SCAN) task reveals the potential of seq2seq recurrent networks in controlled environments \citep{lake2017generalization}. Although models are only exposed to the primitive command before, they are able to understand most composed commands of ``\textit{turn left}''. One assumption is that models study new commands with a primitive from other action sequences containing the basic action it denotes. However, there is still a missing formal exploration to answer the question raised by \citet{lake2017generalization} on page 8 that ``\emph{what are, precisely, the generalization mechanisms that subtend the networks’ success in these experiments}''.

In this work, as a response to the call, we question whether neural networks are indeed deficient or just conventional learning protocols unable to exploit their full potential \citep{csordas2021devil,dankers-etal-2022-transformer}. We revisit the systematic generalization of seq2seq models from a \emph{meaningful learning} perspective \citep{ausubel1963psychology,okebukola1988cognitive,mayer2002rote}. Given the idea that humans are used to memorizing concepts in a relational manner, we hypothesize that the success of the turn-left experiment results from the semantic relationships between old concepts and new ones. For example, in Figure \ref{figure:scan}, a model can understand the meaning of ``\textit{\textbf{turn left and walk} and jump left}'' from ``\textit{\textbf{walk left} and jump left}'' via the semantic link between two concepts (in bold) since both denote to the same action ``LTURN WALK''.

To validate our hypothesis, we reproduce the one-shot compositional generalization by \emph{semantic linking} that exposes semantic relationships through either \emph{inductive learning} or \emph{deductive learning} \citep{hammerly1975deduction,shaffer1989comparison,thornbury1999teach}. On the one hand, by introducing new concepts sharing the same context, we hope the model can capture the underlying semantic connections inductively. On the other hand, by involving a rule-like concept dictionary without specific context information, we hope the model can utilize the general cross-lingual supervised signals as anchor points so as to launch the semantic linking deductively.

In experiments, we treat concepts in the initial data set as primitives and generate variant samples and rules accordingly. Next, we mix them up and construct a seq2seq task after a random split. We repeatedly train and evaluate models but slowly decrease the number of times they see each variant until one-shot learning. We observe there is hardly a performance drop in SCAN for three representative model structures. This evidences that, with semantic linking, even canonical neural networks can generalize systematically to new concepts and compositions. Such observation holds consistently across two more semantic parsing (SP) datasets. The followed sensitivity analysis shows that prior knowledge also takes essential parts. Lastly, as a proof-of-concept, we demonstrate how meaningful learning already benefits models in standard machine translation (MT) and SP. Overall, our contributions\footnote{Code and data are publicly available at \href{https://github.com/ShiningLab/Systematic-Generalization-via-Meaningful-Learning}{GitHub}.} are as follows:
\begin{itemize}[leftmargin=*,noitemsep,topsep=0pt]
    \item We revisit systematic generalization from a meaningful learning perspective by either inductive or deductive semantic linking.
    \item We find that modern seq2seq models can generalize to new concepts and compositions after semantic linking, which empirically answers the question by \citet{lake2017generalization}.
    \item We show in the sensitivity analysis that both semantic linking and prior knowledge play a key role, in line with meaningful learning theory.
    \item We extend to standard MT and SP and demonstrate how meaningful learning already benefits models in solving realistic problems.
\end{itemize}

%% file: 3.meaningful_learning.tex
\section{Meaningful Learning}

In educational psychology, meaningful learning refers to learning new concepts by relating them to old ones \citep{ausubel1963psychology,mayer2002rote}.
In Figure \ref{figure:ml}, intuitively, the utilization of meaningful learning can encourage learners to understand information continuously built on concepts the learners already understand \citep{okebukola1988cognitive}. Following this, we intend to examine models' systematic compositionality by exploring semantic linking that establishes semantic relations between primitives (old concepts) and their variants (new concepts). We propose to spoon-feed semantic knowledge to models for semantic linking in two ways, that is, inductive learning and deductive learning \citep{hammerly1975deduction,shaffer1989comparison,thornbury1999teach}. In this section, we discuss the process of semantic linking and take ``\textit{jump}" from SCAN as an example primitive to illustrate the learning scheme.

\subsection{Semantic links}

We focus on three semantic relationships, namely, \emph{lexical variant}, \emph{co-hyponym}, and \emph{synonym}. Lexical Variant refers to an alternative expression form for the same concept. Co-hyponym is a linguistic term to designate a semantic relation between two group members belonging to the same broader class, where each member is a hyponym and the class is a hypernym \citep{lyons1995linguistic}. Synonym stands for a word, morpheme, or phrase that shares exactly or nearly the same semantics with another one. We provide an example in Figure \ref{figure:ml} and a detailed description in Appendix \ref{appendix:sl}.

\subsection{Inductive learning} \label{sec:il}

Inductive learning is a bottom-up approach from the more specific to the more general. In grammar teaching, inductive learning is a rule-discovery approach starting with the presentation of specific examples from which a general rule can be inferred \citep{thornbury1999teach}. In semantic linking, we propose to introduce variant samples sharing the same context with their primitives during training. The assumption is that models can observe the interchange of primitives and their variants surrounded by the same context in the hope of coming up with a general hypothesis that there is a semantic linking between primitives and their variants \citep{harris1954distributional}. To test the generalization, we design a prompt ``\textit{[concept] twice}'' from a primitive sample ``\textit{jump twice}''. After that, we fill in the concept slot with ``\textit{jump\_0}'' and generate the variant sample ``\textit{jump\_0 twice}''. There is no change from the target side. Finally, by training models on the generated variant sample in combination with prior knowledge (all the other primitive samples), we aim to establish the semantic relationship between ``\textit{jump}'' and ``\textit{jump\_0}'' inductively.

\subsection{Deductive learning} \label{sec:dl}

Deductive Learning, the opposite of inductive learning, is a top-down approach from the more general to the more specific. As a rule-driven approach, teaching in a deductive manner often begins with presenting a general rule followed by specific examples in practice where the rule is applied \citep{thornbury1999teach}. To align with this definition, we intend to do semantic linking deductively by combining a bilingual dictionary that maps primitives and their variants to the same in the target domain. This additional dictionary, hence, mixes the original training task with word translation \citep{mikolov2013exploiting}. Without any specific context, we hope the model can utilize the general cross-lingual supervised signals as anchor points so as to launch the semantic linking. We want to point out that deductive learning is partially different from \emph{deductive reasoning}. Although there is an overlap, it is not necessary for the former to extract rules from observations like the inference conducted by the latter. In this work, we care more about the learning outcomes, rather than the reasoning process, through empirical evaluations. In practice, given the same example above, we directly make use of primitive ``\textit{jump}'' and its variant ``\textit{jump\_0}'' as the source sequences, as well as the action ``JUMP'' as their identical target sequences. Words and phrases can be treated as text sequences of relatively short length. By exposing both the primitive rule ``\textit{jump}'' $\rightarrow$ ``JUMP'' and the variants rule ``\textit{jump\_0}''$\rightarrow$ ``JUMP'' during training, we aim to build the semantic connections between ``\textit{jump}'' and ``\textit{jump\_0}'' deductively.

%% file: 4.systematic_generalization.tex
\begin{figure*}[t]
    \begin{center}
        \includegraphics[width=\textwidth]{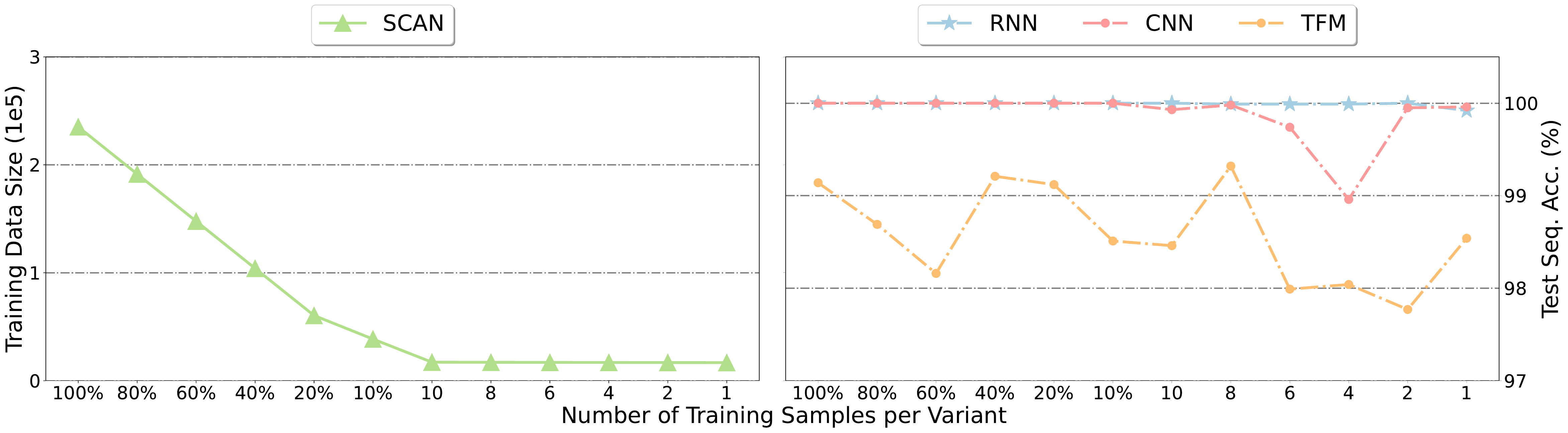}
        \vspace{-8mm}
    \end{center}
    \caption{Experiments on SCAN expressing the total training size (left) and the test sequence accuracy (right) when the number of training samples per variant decreases from the complete set (100\%) to a single sample (1).}
\label{figure:exp0}
\vspace{-2mm}
\end{figure*}

\section{Systematic Generalization}

The following section specifies the setup and outcome of the experiments. We first employ SCAN as the initial testbed to reproduce the one-shot generalization conditioned on the semantic linking. Then, we examine neural networks' potential to achieve this on SCAN and two real-world tasks of SP, followed by a sensitivity analysis.

\subsection{Datasets}

Some suggest SCAN is not enough to fully verify compositionality \citep{bastings-etal-2018-jump,keysers2020measuring,dankers-etal-2022-paradox}. Thus, we introduce GEO and ADV generated respectively from real SP datasets: Geography and Advising.\footnote{\url{github.com/jkkummerfeld/text2sql-data}} Example inputs and outputs can be found in Table \ref{table:data}.\\
\textbf{SCAN} \citep{lake2017generalization} is a diagnostic dataset proposed to investigate neural networks' compositionality.\footnote{\url{github.com/brendenlake/SCAN}} It includes 20,910 pairs of commands to their instructed actions such as the example in Figure \ref{figure:scan}. We select \{``\textit{jump}", ``\textit{look}", ``\textit{run}", ``\textit{walk}" \} as 4 primitives to be in line with previous works. We focus on lexical variants and create them by adding a suffix that consists of an underline and a unique number. We control the size of the variants set by setting the upper limit of this number. An example variant of ``\textit{jump}” is ``\textit{jump\_0}" and both mean the same action ``JUMP". \\
\textbf{Geography} is a common SP dataset \citep{data-geography-original,data-atis-geography-scholar}, containing 880 examples of queries paired with corresponding expressions. It is later formatted to SQL language with variables in the target sequences \citep{data-sql-advising}. \textbf{GEO} is generated from Geography. We regard 4 of 9 annotated variables as hypernyms and keep them as they are in SQL sequences. The other variables are restored by entities from the source sequence accordingly. As a result, the overall data size is 618 after processing. We can make use of the ``is-a" hypernymy relation for semantic linking. Specifically, we select \{``\textit{new york city}", ``\textit{mississippi rivier}", ``\textit{dc}", ``\textit{dover}" \} as 4 primitives\footnote{We randomly select 4 primitives from GEO and ADV to align with SCAN.} with their variants consisting of entities as co-hyponyms sharing the same variable group with primitives. An example variant of ``\textit{new york city}” is ``\textit{houston city}" and both are in the same variable group ``CITY\_NAME''. \\
\textbf{Advising} includes 4,570 questions on course information paired with SQL queries \citep{data-sql-advising}. \textbf{ADV} is generated from Advising. We treat 4 of 26 annotated variables as hypernyms. Precisely, we select \{ ``\textit{a history of american film}", ``\textit{aaron magid}", ``\textit{aaptis}", ``\textit{100}" \} as 4 primitives with their variants as co-hyponyms sharing the same variables. For instance, ``\textit{advanced at ai techniques}" is a co-hyponym of ``\textit{a history of american film}" sharing the same variable ``TOPIC".

\begin{table*}[t]
  \resizebox{\textwidth}{!}{
  \centering
  \begin{tabular}{llccccccccccccccc}
    \toprule
    & & \multicolumn{5}{c}{SCAN} & \multicolumn{5}{c}{GEO} & \multicolumn{5}{c}{ADV} \\
    \cmidrule(r){3-7}
    \cmidrule(r){8-12}
    \cmidrule(r){13-17}
    \multicolumn{2}{l}{\textbf{Data}} & \multicolumn{3}{c}{Exp. IL} & \multicolumn{2}{c}{Exp. DL} & \multicolumn{3}{c}{Exp. IL} & \multicolumn{2}{c}{Exp. DL} & \multicolumn{3}{c}{Exp. IL} & \multicolumn{2}{c}{Exp. DL} \\
    \cmidrule(r){3-5}
    \cmidrule(r){6-7}
    \cmidrule(r){8-10}
    \cmidrule(r){11-12}
    \cmidrule(r){13-15}
    \cmidrule(r){16-17}
    & & Sta. & Dif. & Cha. & Sta. & Dif. & Sta. & Dif. & Cha. & Sta. & Dif. & Sta. & Dif. & Cha. & Sta. & Dif. \\
    \midrule
    \multicolumn{2}{l}{\textbf{Train Size}} & 20946 & 20942 & 20928 & 20950 & 20946 & 724 & 720 & 711 & 728 & 724 & 6038 & 6034 & 5969 & 6040 & 6036 \\
    \multicolumn{2}{l}{\textbf{Test Size}}  & 308240 & 308240 & 308240 & 308240 & 308240 & 21350 & 21350 & 21350 & 21350 & 21350 & 107614 & 107614 & 107614 & 107614 & 107614 \\
    \bottomrule
  \end{tabular}
  }
  \vspace{-2mm}
  \caption{Dataset statistics for inductive learning (IL) and deductive learning (DL) across Standard (Sta.), Difficult (Dif.), and Challenging (Cha.) in Section \ref{sec:sli}.}
  \label{table:data_il&dl}
\vspace{-2mm}
\end{table*}

\subsection{Models and experimental setup}

\textbf{Models.} After testing many adapted versions, we employ three dominant model candidates, that is, RNN, CNN, and TFM. In terms of RNN, we reproduce bi-directional recurrent networks \citep{schuster1997bidirectional} with long short-term memory units \citep{hochreiter1997long} and an attention mechanism \citep{bahdanau2014neural}. We follow the convolutional seq2seq architecture presented by \citet{gehring2017convolutional} with regard to CNN and the attention-based structure proposed by \citet{NIPS2017_7181} in the case of TFM. More details are in Appendix \ref{appendix:model}. \\
\textbf{Training.} We apply the mini-batch strategy to sample 128 sequence pairs for each training step. We use Adam optimizer \citep{DBLP:journals/corr/KingmaB14} with an $\ell_2$ gradient clipping of $5.0$ \citep{10.5555/3042817.3043083} and a learning rate of 1$e^{-4}$. We freeze the maximum training epoch at 320 for CNN and 640 for RNN and TFM. To prevent uncontrolled interference, we train all models from scratch instead of fine-tuning \citep{devlin-etal-2019-bert}. For the same reason, we break words by whitespace tokenization rather than subword modeling. So, we can guarantee that words are treated separately as distinct tokens with completely different embeddings.\\
\textbf{Evaluation.} Token and sequence accuracy serve as two primary metrics. The former allows partial errors in a sequence, while the latter strictly does not. Every reported number, along with the standard deviation, is the mean of five runs.

\subsection{Experiment: meaningful learning}

Thanks to their incredible algebraic compositionality \citep{chomsky1957syntactic}, humans can effectively capture the underlying semantic connections between new and old concepts and generalize the prior knowledge to novel combinations by meaningful learning \citep{ausubel1963psychology}. To investigate the extent to which models can do the same, we probe the models' compositionality by introducing semantic linking. It is reasonable to illustrate the function of semantic linking through an ablation study, while its missing will lead to an out-of-vocabulary (OOV) issue since there will be no sample to expose variants during training. Replacing variants with other tokens (e.g., “\textit{[unk]}”) goes against our intent to investigate the generalization from primitives to their variants. It also leads to an unfair comparison, where all the variants, for example, go to the same unknown token and cause poor test accuracy. Instead, we gradually remove training samples for each variant until the one-shot learning scenario. We hope to observe the presence of models' meaningful learning by measuring the corresponding performance loss. \\
\textbf{Experimental setup.} Following section \ref{sec:il}, we make use of 40 variants for 4 primitives and produce a total of 329,190 samples, including both primitive and variant samples. We randomly split them into a training set (80\%) and a test set (20\%). The training set is further processed to remove samples having multiple variants to ensure that each variant occurs only once in each sample. Eventually, the training set contains 235,002 samples. Models directly trained on this full dataset serve as baselines. Then, to format a gradual transition from baselines to the meaningful learning, we train the same models on various datasets with a decreasing number of augmented samples for each variant until the one-shot learning setting. Besides, we use the variant rule ``\textit{jump\_0}'' $\rightarrow$ ``JUMP'' as the only training sample for ``\textit{jump\_0}'' in the end as a case of our deductive learning introduced in Section \ref{sec:dl} and consider the rest as our inductive learning. \\
\textbf{Results.} As elaborated in Figure \ref{figure:exp0}, the solid line (SCAN) in green denotes the total training data size against the decreasing number of training samples per variant. The dashed line in other colors denotes the test sequence accuracy against the same horizontal axis. RNN has no significant performance drop when the training size is reduced from 100\% to 1. It still achieves 99.92\% test sequence accuracy when there is only one training sample for each variant. The same happens for CNN and TFM. Despite a slight fluctuation, they keep the results almost consistent regardless of whether the number of training variant samples is all or 1. It is not necessary to augment the training set nearly 14 times from 16,736 to 235,002 to cover all the possible variant compositions. The participation of a single sample is able to launch semantic linking via either inductive learning (a variant sample) or deductive learning (a variant rule), thus enabling models to achieve one-shot generalization. We put two plots in one figure to emphasize such a surprising observation through the strong contrast.

\begin{table*}[t]
  \begin{center}
  \resizebox{\textwidth}{!}{
  \begin{tabular}{llcccccc}
    \toprule
    \multirow{2}{*}{\textbf{Data}} & \multirow{2}{*}{\textbf{Model}} & \multicolumn{3}{c}{\textbf{Token Acc.\%}} & \multicolumn{3}{c}{\textbf{Seq. Acc.\%}} \\
    \cmidrule(r){3-5}
    \cmidrule(r){6-8}
    & & \textbf{Standard} & \textbf{Difficult} & \textbf{Challenging} & \textbf{Standard} & \textbf{Difficult} & \textbf{Challenging} \\
    \midrule
     & RNN & $99.99\pm0.03$ & $99.89\pm0.19$ & $99.96\pm0.02$ & $99.95\pm0.08$ & $99.85\pm0.08$ & $99.80\pm0.31$ \\
    SCAN & CNN & $99.96\pm0.08$ & $99.76\pm0.54$ & $98.89\pm2.44$ & $99.85\pm0.34$ & $99.52\pm1.07$ & $97.57\pm5.24$ \\
     & TFM & $98.91\pm0.78$ & $98.90\pm1.10$ & $98.76\pm0.85$ & $97.35\pm1.62$ & $96.86\pm2.64$ & $96.38\pm2.81$ \\
    \midrule
     & RNN & $75.71\pm8.42$ & $75.69\pm6.12$ & $73.46\pm3.05$ & $44.95\pm14.69$ & $43.27\pm13.47$ & $36.77\pm5.60$ \\
    GEO & CNN & $87.99\pm2.67$ & $79.51\pm6.03$ & $77.40\pm2.48$ & $69.46\pm5.78$ & $51.20\pm8.64$ & $48.58\pm3.40$ \\
     & TFM & $75.37\pm7.84$ & $75.11\pm4.88$ & $68.41\pm4.76$ & $45.93\pm12.42$ & $44.59\pm9.76$ & $36.93\pm7.47$ \\
    \midrule
     & RNN & $58.61\pm6.18$ & $59.74\pm5.67$ & $58.11\pm5.82$ & $36.18\pm5.75$ & $35.69\pm6.05$ & $35.45\pm6.69$ \\
    ADV & CNN & $57.83\pm7.55$ & $54.05\pm5.74$ & $53.66\pm2.57$ & $45.08\pm9.32$ & $42.14\pm6.90$ & $41.37\pm4.04$ \\
     & TFM & $53.43\pm2.80$ & $51.51\pm4.50$ & $49.17\pm2.58$ & $42.59\pm3.65$ & $41.28\pm4.35$ & $38.88\pm2.68$ \\
    \bottomrule
  \end{tabular}
  }
  \end{center}
  \vspace{-2mm}
  \caption{Evaluation results over RNN, CNN, and TFM on SCAN, GEO, and ADV across Standard, Difficult, and Challenging in Section \ref{subsec:il}.}
  \label{table:il}
  \vspace{-4mm}
\end{table*}

\subsection{Experiment: semantic linking injection} \label{sec:sli}

The following two experiments evaluate models' systematic generalization, particularly for prior knowledge and semantic linking. A sliding scale of difficulty is carefully designed by weakening these two factors according to the assumption that the greater the difficulty, the more compositional skills are required. We further validate our findings on GEO and ADV. We use the same evaluation protocol across different datasets in this section.

Taking the base dataset as prior knowledge, we replace the primitives in source sequences with their variants to generate novel compositions, as introduced in Section \ref{sec:il}. So far, the produced variant samples are not in the training set but in the test set. Hence, variants exist as OOV now. Then, we either incorporate one variant sample to introduce variants in training inductively or one variant rule to do so deductively. In the one-shot learning scenario, we ensure each variant only has a single sample and appears only once during training. For convenience, we keep the same settings for each primitive to have 10 variants in SCAN and a full variant set in GEO (e.g., 39 variants for ``\textit{new york city}"). It is noted in ADV that we randomly sample 5 variants for each primitive so that we cover all the variants with an appropriate test size.

\subsubsection{Inductive learning} \label{subsec:il}

\textbf{Experimental setup.} We increase the difficulty by excluding primitive samples from the training set. It is worth noting that models have to generalize to not only new concepts but also their new compositions with a higher level of difficulty.
\begin{itemize}[leftmargin=*,noitemsep,topsep=0pt]
    \item \textbf{Standard}: Models are trained on prior knowledge and one variant sample per variant.
    \item \textbf{Difficult}: We remove from the prior knowledge primitive samples sharing the same context with their variant samples. For example, we remove ``\textit{jump twice}'' due to ``\textit{jump\_0 twice}'', and thus models have to generalize to ``\textit{jump\_0 twice}'' without seeing ``\textit{jump twice}''.
    \item \textbf{Challenging:} We also exclude from the prior knowledge primitive samples of the same length as their variant samples. For instance, models have to reproduce the same generalization to ``\textit{jump\_0 twice}'' without seeing primitive samples of length 2, including ``\textit{jump twice}'', ``\textit{jump right}'', ``\textit{jump left}", to name a few.\footnote{We remove samples that will not lead to unknown tokens.}
\end{itemize}
\begin{table}[t]
  \begin{center}
  \resizebox{\linewidth}{!}{
  \begin{tabular}{llcccc}
    \toprule
    \multirow{2}{*}{\textbf{Data}} & \multirow{2}{*}{\textbf{Model}} & \multicolumn{2}{c}{\textbf{Token Acc.\%}} & \multicolumn{2}{c}{\textbf{Seq. Acc.\%}} \\
    \cmidrule(r){3-4}
    \cmidrule(r){5-6}
    & & \textbf{Standard} & \textbf{Difficult} & \textbf{Standard} & \textbf{Difficult} \\
    \midrule
    & RNN & $99.48\pm0.71$ & $98.70\pm0.92$ & $98.27\pm2.38$ & $95.39\pm2.72$ \\
    SCAN & CNN & $99.99\pm0.01$ & $98.59\pm3.10$ & $99.96\pm0.03$ & $96.66\pm7.27$ \\
    & TFM & $96.90\pm1.78$ & $96.68\pm2.21$ & $91.94\pm4.04$ & $91.26\pm5.80$ \\
    \midrule
    & RNN & $54.44\pm7.15$ & $39.71\pm18.38$ & $13.61\pm7.08$ & $7.76\pm5.34$ \\
    GEO & CNN & $41.86\pm3.38$ & $41.07\pm7.48$ & $4.85\pm4.66$ & $4.04\pm2.18$ \\
    & TFM & $67.02\pm6.91$ & $65.97\pm5.17$ & $36.38\pm10.08$ & $31.57\pm7.42$ \\
    \midrule
    & RNN & $36.50\pm7.66$ & $36.42\pm7.39$ & $12.84\pm4.31$ & $12.66\pm5.19$ \\
    ADV & CNN & $43.51\pm11.31$ & $35.34\pm14.68$ & $32.33\pm12.93$ & $23.58\pm16.04$ \\
    & TFM & $56.82\pm3.79$ & $53.33\pm3.85$ & $47.43\pm3.71$ & $43.24\pm5.14$ \\
    \bottomrule
  \end{tabular}
  }
  \end{center}
  \vspace{-2mm}
  \caption{Evaluation results over RNN, CNN, and TFM on SCAN, GEO, and ADV across Standard and Difficult in Section \ref{subsec:dl}.}
  \label{table:dl}
\vspace{-4mm}
\end{table}
\textbf{SCAN.} What stands out in Table \ref{table:il} is an excellent one-shot generalization for all three networks. The participation of variant samples induces a near-perfect generalization. Even the worst results obtained by TFM in Challenging are around $98.76\%$ and $96.38\%$ in terms of token and sequence accuracy. The outcomes confirm that networks can inductively learn semantic relations from the context after semantic linking. The disappearance of training samples in Difficult and Challenging causes a performance drop. This is well in line with the widely accepted belief in meaningful learning theory that prior knowledge matters to generalization. \\
\textbf{GEO \& ADV.} The more apparent changes in metrics again verify that prior knowledge is essential. Either excluding primitive samples containing the same context or those of the same sequence length can produce a steep fall in the generalization. On GEO, CNN can lose an absolute sequence accuracy of $18.26\%$ from Standard to Difficult, and that for TFM drops $7.66\%$. This upholds our argument that generalization via meaningful learning is inseparable from sufficient prior knowledge. The overall decline in performance can be attributed to the switch from toy sets to actual datasets since both GEO and ADV own a much more complex encoding and decoding space than SCAN. Therefore, we conclude that both prior knowledge and semantic linking exert powerful effects upon the potential of models to generalize systematically.

\subsubsection{Deductive learning} \label{subsec:dl}

\textbf{Experimental setup.} We increase the difficulty of compositional learning by excluding primitive rules from the training set as follows:
\begin{itemize}[leftmargin=*,noitemsep,topsep=0pt]
    \item \textbf{Standard}: Models are trained on the prior knowledge, primitive rules, and variant rules.
    \item \textbf{Difficult}: We remove primitive rules from the training set. Consequently, semantic links are weakened and depend on variant rules only.
\end{itemize}
\textbf{SCAN.} By incorporating deductive semantic linking, all three networks attain satisfying compositional generalization as shown in Table \ref{table:dl}. CNN achieves the highest $99.96\%$ in Standard, while TFM takes the lowest $91.26\%$ in Difficult with regard to sequence accuracy. We can see a consistent decline in accuracy when we undermine the semantic linking by removing primitive rules from the training set. The most significant sequence accuracy drop of $3.3\%$ comes from CNN when the difficulty upgrades. However, in Difficult, even the lowest one is impressive as there is only one variant rule to introduce each variant during training. \\
\textbf{GEO \& ADV.} There is a persistent performance loss because of the absence of primitive rules from the training set across models. Concretely in GEO, the grade of CNN declines from $32.33\%$ in Standard to $23.58\%$ in Difficult in terms of sequence accuracy. The causal role of semantic linking is also demonstrated by varying the difficulty. The difference between Standard and Difficult indicated that either concept rules and just variant rules can connect primitives with their variants semantically, though the former is better than the latter. Moreover, models appear to realize systematic generalization better in an inductive way. By comparing Table \ref{table:il} with Table \ref{table:dl}, we find that current black-box neural nets are more capable of exploring patterns from specific samples with context information rather than understanding knowledge from general rules in our experiments. This sheds light on why current machine learning is still highly data-driven and can hardly break through the bottleneck to conduct advanced logic reasoning as human beings.

\begin{figure}[t]
    \begin{center}
        \includegraphics[width=\linewidth]{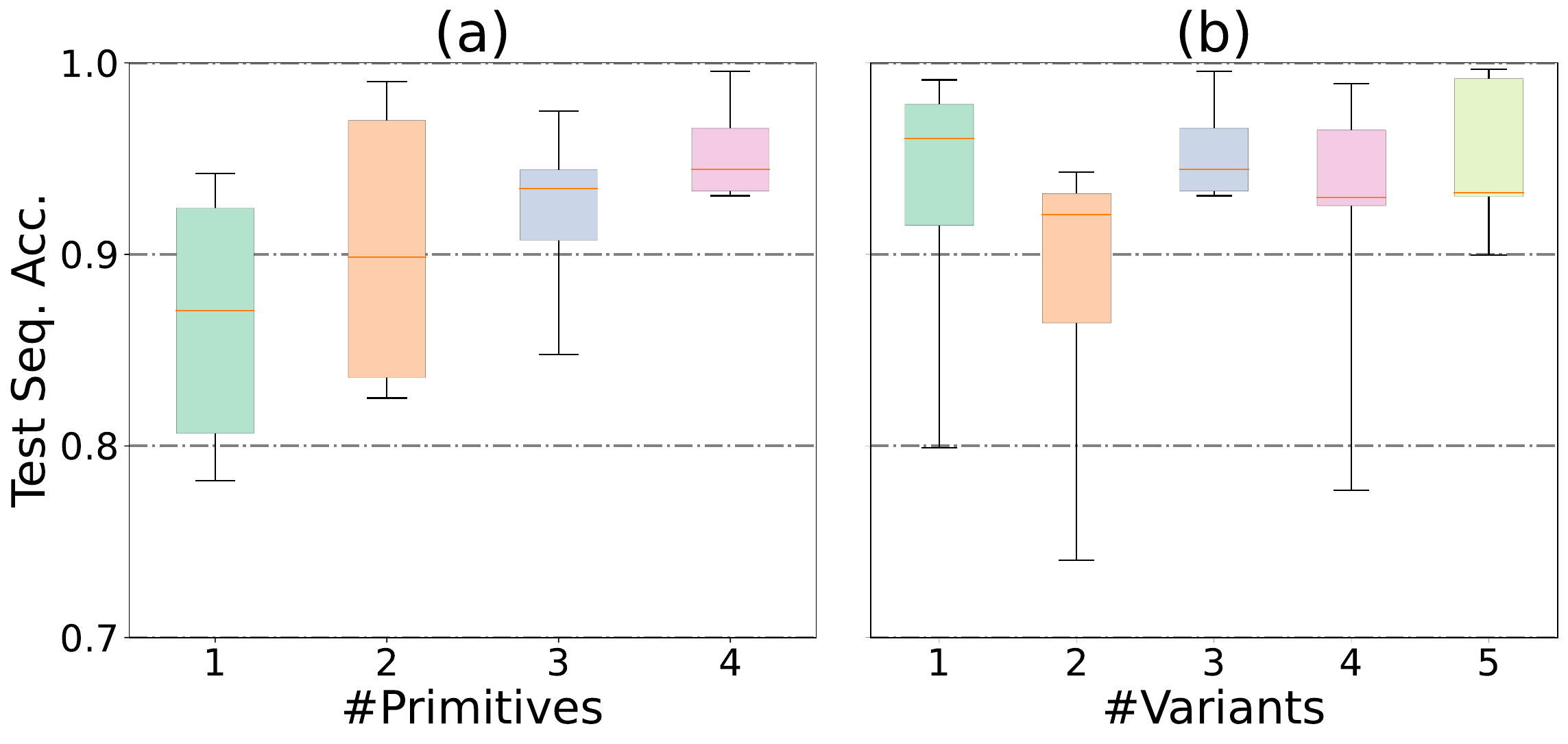}
        \vspace{-8mm}
    \end{center}
\caption{Experiments over RNN on SCAN with varying \#primitives (a) and \#variants (b).}
\label{figure:sa}
\vspace{-2mm}
\end{figure}

\subsection{Sensitivity analysis} \label{sec:sa}

\begin{table*}[t]
  \resizebox{\textwidth}{!}{
  \centering
  \begin{tabular}{lcccccccc}
    \toprule
     & \multicolumn{4}{c}{\textbf{IWSLT'14}} & \multicolumn{4}{c}{\textbf{IWSLT'15}} \\
    \cmidrule(r){2-5}
    \cmidrule(r){6-9}
    \textbf{Model} & \multicolumn{2}{c}{\textbf{En-De}} & \multicolumn{2}{c}{\textbf{De-En}} & \multicolumn{2}{c}{\textbf{En-Fr}} & \multicolumn{2}{c}{\textbf{Fr-En}} \\
    \cmidrule(r){2-3}
    \cmidrule(r){4-5}
    \cmidrule(r){6-7}
    \cmidrule(r){8-9}
     & \textbf{BLEU} & \textbf{SacreBLEU} & \textbf{BLEU} & \textbf{SacreBLEU} & \textbf{BLEU} & \textbf{SacreBLEU}  & \textbf{BLEU} & \textbf{SacreBLEU} \\
    \midrule
    \multicolumn{9}{l}{\textbf{Baselines}} \\
    LSTM \citep{luong-etal-2015-effective} & 24.98 & 24.88 & 30.18 & 32.62 & 38.06 & 42.93 & 37.34 & 39.36 \\
    Transformer \citep{NIPS2017_7181} & 28.95 & 28.85 & 35.24 & 37.60 & 41.82 & 46.41 & 40.45 & 42.61 \\
    Dynamic Conv. \citep{wu2018pay} & 27.39 & 27.28 & 33.33 & 35.54 & 40.41 & 45.32 & 39.61 & 41.42 \\
    \midrule
    \multicolumn{9}{l}{\textbf{\texttt{\Large+}Vocabulary Augmentation}} \\
    LSTM \citep{luong-etal-2015-effective} & 25.35$\uparrow_{0.37}$ & 25.38$\uparrow_{0.50}$ & 30.99$\uparrow_{0.81}$ & 33.63$\uparrow_{1.01}$ & 38.32$\uparrow_{0.26}$ & 43.30$\uparrow_{0.37}$ & 37.77$\uparrow_{0.43}$ & 39.83$\uparrow_{0.47}$ \\
    Transformer \citep{NIPS2017_7181} & 29.40$\uparrow_{0.45}$ & 29.29$\uparrow_{0.44}$ & 35.72$\uparrow_{0.48}$ & 38.07$\uparrow_{0.47}$ & 42.19$\uparrow_{0.37}$ & 46.68$\uparrow_{0.27}$ & 41.04$\uparrow_{0.59}$ & 43.15$\uparrow_{0.54}$ \\
    Dynamic Conv. \citep{wu2018pay} & 27.60$\uparrow_{0.21}$ & 27.50$\uparrow_{0.22}$ & 33.62$\uparrow_{0.29}$ & 36.00$\uparrow_{0.46}$ & 40.87$\uparrow_{0.46}$ & 45.95$\uparrow_{0.63}$ & 39.95$\uparrow_{0.34}$ & 41.86$\uparrow_{0.44}$ \\
    \bottomrule
  \end{tabular}}
  \vspace{-2mm}
  \caption{Evaluation results over LSTM, Transformer, and Dynamic Conv. on IWSLT'14 En-De (English-German) and De-En, IWSLT'15 En-Fr (English-French) and Fr-En translations.}
  \label{table:mt}
\vspace{-4mm}
\end{table*}

Regarding deductive learning, we conduct sensitivity analysis with a varying number of primitives (\#primitives) from \{1,2,3,4\} and that of variants per primitive (\#variants) from \{1,5,10,15,20\} over RNN on SCAN. The experimental setup is borrowed from Standard in Section \ref{subsec:dl}. \\
\textbf{Impact of \#primitives.} In Figure \ref{figure:sa} (a), the generalization performance improves w.r.t. accuracy boosting and variance reduction when \#primitives grows simultaneously. This is counter-intuitive as we thought primitive rules should work independently. A potential reason is semantic linking built by various \emph{independent} primitive rules can profit each other to trigger a more robust and stable generalization. For example, ``\textit{jump}'' $\rightarrow$ ``JUMP'' and ``\textit{look}'' $\rightarrow$ ``LOOK'' may separate them from the context such as ``\textit{\textbf{jump} right}'' and ``\textit{\textbf{look} right}''. So, ``\textit{[concept] right}'' functions as a compositional rule shared among primitive samples and finally encourages models to generalize more effectively. \\
\textbf{Impact of \#variants.} As presented in Figure \ref{figure:sa} (b), RNN generalizes consistently well when \#variants goes up. Therefore, we report that the generalization among variants of the same primitive has a certain degree of independence within a reasonable range (e.g., \#variants $\leq 20$).

%% file: 5.discussion.tex
\section{From SCAN to Real Data}

Thus far, we have argued the feasibility of systematic generalization activated by semantic linking. We move on to discuss how it already benefits machines in solving real problems. Many recent papers propose to improve systematic generalization by techniques such as data augmentation \citep{andreas-2020-good,rek2021learning} and meta-learning \citep{NEURIPS2019_f4d0e2e7,conklin-etal-2021-meta}. The success is reasonable given our findings. Replacing fragments in real training samples with others that share similar contexts is supported by our inductive learning. We have demonstrated that similar context information can help establish the semantic links between new concepts and old ones, thus enabling models to generalize compositionally. By considering concepts as pointers in the memory, meta-learning equips models with memory loading to make connections between new and old concepts as semantic linking. The utility of similar unsupervised techniques \citep{xie2019unsupervised} in both compositional generalization and real tasks can be attributed to inductive learning as well. Besides, our sensitivity analysis in Section \ref{sec:sa} shows that adding seemingly independent primitive samples or rules can also improve the generalization, which has been further validated recently \citep{auersperger-pecina-2021-solving,patel-etal-2022-revisiting}.

In addition to inductive-based methods, some works \citep{mikolov2013exploiting,arthur-etal-2016-incorporating, nag2020incorporating}, incorporating bilingual dictionaries in low-resource MT, can fall in the field of deductive-based ones. As a proof-of-concept, we reproduce the word-to-word augmentation, or called deductive learning in this work, by training models on not only the base training set but also concept rules. Intuitively, we wonder to which extent deductive semantic linking can promote models' performance in MT (IWSLT'14 and IWSLT'15) and SP (Geography and Advising). We report the evaluation results in Table \ref{table:mt} and Table \ref{table:sp}. Details of models and data can be found in Appendix \ref{appendix:model} and Appendix \ref{appendix:data}.

\begin{table*}[t]
  \resizebox{\textwidth}{!}{
  \centering
  \begin{tabular}{lcccccccc}
    \toprule
     & \multicolumn{4}{c}{\textbf{Geography}} & \multicolumn{4}{c}{\textbf{Advising}} \\
    \cmidrule(r){2-5}
    \cmidrule(r){6-9}
    \textbf{Model} & \multicolumn{2}{c}{\textbf{Train}} & \multicolumn{2}{c}{\textbf{Test}} & \multicolumn{2}{c}{\textbf{Train}} & \multicolumn{2}{c}{\textbf{Test}} \\
    \cmidrule(r){2-3}
    \cmidrule(r){4-5}
    \cmidrule(r){6-7}
    \cmidrule(r){8-9}
     & \textbf{Token Acc.\%} & \textbf{Seq. Acc.\%} & \textbf{Token Acc.\%} & \textbf{Seq. Acc.\%} & \textbf{Token Acc.\%} & \textbf{Seq. Acc.\%} & \textbf{Token Acc.\%} & \textbf{Seq. Acc.\%} \\
    \midrule
    \multicolumn{8}{l}{\textbf{Baselines}} \\
    RNN & 89.05 & 17.39 & 69.81 & 9.68 & 92.22 & 3.64 & 60.41 & 6.11 \\
    CNN & 98.45 & 70.74 & 78.44 & 55.91 & 99.74 & 81.62 & 81.74 & 51.13 \\
    TFM & 99.45 & 84.95 & 80.24 & 49.82 & 99.68 & 76.90 & 78.51 & 29.67 \\
    \midrule
    \multicolumn{8}{l}{\textbf{\texttt{\Large+}Entity Augmentation}} \\
    RNN & 87.47 & 29.96 & 72.39$\uparrow_{2.58}$ & 15.05$\uparrow_{5.37}$ & 88.82 & 30.97 & 71.17$\uparrow_{10.76}$ & 16.06$\uparrow_{9.95}$ \\
    CNN & 97.54 & 76.03 & 80.32$\uparrow_{1.88}$ & 60.93$\uparrow_{5.02}$ & 99.65 & 87.01 & 84.50$\uparrow_{2.76}$ & 56.02$\uparrow_{4.89}$ \\
    TFM & 99.30 & 85.73 & 81.09$\uparrow_{0.85}$ & 54.84$\uparrow_{5.02}$ & 99.57 & 86.94 & 84.26$\uparrow_{5.75}$ & 35.08$\uparrow_{5.41}$ \\
    \bottomrule
  \end{tabular}}
  \vspace{-2mm}
  \caption{Evaluation results over RNN, CNN, and TFM on Geography and Advising.}
  \label{table:sp}
\vspace{-4mm}
\end{table*}

\subsection{Machine translation}
\textbf{Setup.}
We evaluate our approach on IWSLT'14 \cite{cettolo2014report} English-German (En-De) and German-English (De-En), IWSLT'15 \cite{Cettolo2015TheI2} English-French (En-Fr) and French-English (Fr-En) translation tasks. We follow the standard evaluation protocol \cite{ott2019fairseq} that keeps the original training set and validation set but combines multiple previous test sets for final evaluation. The test set of IWSLT'14 consists of IWSLT14.TED.dev\{2010, 2012\} and IWSLT14.TED.tst\{2010, 2011, 2012\}. That of IWSLT'15 includes IWSLT15.TED.tst\{2014, 2015\} \cite{ott2019fairseq}. We apply BPE with 10K tokens for all tasks and report both BLEU \cite{papineni2002bleu} and SacreBLEU \cite{post-2018-call} scores for three baselines: LSTM \cite{luong-etal-2015-effective}, Transformer \cite{NIPS2017_7181}, and Dynamic Conv. \cite{wu2018pay} in comparsion with same structures augmented by our method.
\\
\textbf{Vocabulary augmentation.}
We introduce concept rules as \emph{vocabulary augmentation} in MT. The semantic links between primitives and their variants can be built upon the synonymous relations between tokens such as ``\textit{heavily populated}" and ``\textit{populous}". From this, the source words paired with translated ones can be regarded as concept rules. It is noted that such relationships are reversible as shown in Figure \ref{figure:ml}, so a primitive can be a variant of the other primitive as well. In practice, we collect a dictionary of tokens in the source language and feed them to the Google Translation\footnote{\url{cloud.google.com/translate}} so as to obtain a token map from the source language to the target one. The same operation can be repeated from the target language to the source one. Two dictionaries are combined into one with duplicates removed. Consequently, we get 144,874 token-level samples as a training supplementary for IWSLT'14 En-De and De-En, and 110,099 for IWSLT'15 En-Fr and Fr-En, which leads to a total of 305,113 training samples for IWSLT'14 En-De and De-En and 315,671 for IWSLT'15 En-Fr and Fr-En after such vocabulary augmentation.
\\
\textbf{Results.}
From Table \ref{table:mt}, we observe a consistent improvement in both BLEU and SacreBLEU over all baselines after vocabulary augmentation, particularly up to 1 in SacreBlEU. The additional synonym pairs not only construct the semantic linking between tokens in two languages explicitly, but also create a complicated semantic linking network implicitly because of synonyms within the single language and the transitivity nature of synonym relation. Our experiments prove that semantic linking, which allows models to generalize systematically, can be beneficial for improving MT performance.

\subsection{Semantic parsing}
\textbf{Setup.}
We evaluate our method on two SP benchmarks, Geography, and Advising. We train the same models (i.e., RNN, CNN, and TFM) as we analyzed before without further hyperparameter tuning. There are some changes for CNN, where the learning rate is 5$e^{-4}$ in Geography, and the maximum sequence length for the decoder position embedding is 312 in Advising. We split $10\%$ training samples as the validation set to find the converged epoch and then add it back to the training set for the final report.
\\
\textbf{Entity augmentation.}
We introduce concept rules as \emph{entity augmentation} in SP. The semantic links are established among co-hyponyms. We consider a variable as a hypernym for its values. By that, entities belonging to the same variable are co-hyponyms. Thus, we can regard entity values as primitives and the translations from primitives (e.g., ``\textit{new york city}'') to their variables (e.g., ``CITY\_NAME'') as primitive rules. To be specific, We construct entity dictionaries by collecting entities such as ``\textit{new york city}''. They are translated to themselves since they do not change from the source natural language to the target SQL. For a fair comparison, a token from this extra dataset will be marked as a unique unknown mark, ``\textit{[unk]}", if it does not exist in the original base training set. After that, we have a map of 103 entity translations for  Geography and 1846 for Advising, resulting in a training size change from 701 to 804 for Geography and from 3814 to 5660 for Advising.
\\
\textbf{Results.}
As elaborated in Table \ref{table:sp}, all three networks can achieve better performance in terms of both accuracy and variance. A $10.76\%$ token accuracy and $9.95\%$ sequence accuracy boosting are observed from RNN on Advising after such entity augmentation. The results suggest that models can learn semantic linking or be more familiar with similar contexts from those primitive rules in a deductive way to enhance model systematic generalization and finally lead to better outcomes. 

%% file: 6.conclusion.tex
\section{Conclusion}

We revisit systematic generalization from a meaningful learning perspective. According to the theory, we conduct semantic linking to expose semantic relations between new and old concepts via either inductive learning or deductive learning. Experimental results on SCAN, GEO, and ADV support that seq2seq neural networks, as a class of modern machine learning methods, can behave systematically after semantic linking. Testing with various difficulties indicates that both semantic linking and prior knowledge are two essential factors in such generalization, in agreement with what humans do in meaningful learning. Finally, we group recent methods in either the inductive-based or deductive-based category, followed by a proof-of-concept, to highlight the already-existing advantages of meaningful learning in applications such as machine translation and semantic parsing.

We want to underline that, to the best of our knowledge, this work is the first one exploring the optimistic results observed by \citet{lake2017generalization}. Our positive findings oppose the recent prevailing view that neural networks appear inherently ineffective in such cognitive capacity, thus confirming the mixed picture. By rationalizing recent findings from a meaningful learning perspective, we hope to encourage followers to interpret the exceptional generalization ability through the connection between neural nets and human cognition.

%% file: 7.others.tex
\section*{Limitations}
We establish semantic relationships between primitives and their variants by either inductive or deductive learning. The incorporation of both learning skills is worth exploring further. We primarily utilize data augmentation techniques to expose the semantic information to models. Apart from that, there should be many other methods to achieve the same goal. Which method is most appropriate to realize semantic linking remains an open topic. Meanwhile, the application of meaningful learning to promote systematic generalization in practice (e.g., MT and SP) could have been expanded.

\section*{Acknowledgements}
We gratefully appreciate Yewen Pu, Guan (Royal) Wang, Yichen Gong, Rong Zhang, and Hui Xue for sharing their pearls of wisdom. We also would like to express our special thanks of graitude to Yingying Huo for the support, as well as BlackboxNLP anonymous reviewers for their constructive feedback. This work was supported by Shining Lab, Learnable, Inc., and Alibaba Group.

%% file: 8.appendix.tex
\section{Semantic Links} \label{appendix:sl}

\begin{table*}[htp!]
  \resizebox{\textwidth}{!}{
  \centering
  \begin{tabular}{lll}
    \toprule
    \textbf{Data} & \textbf{Sequence} \\
    \midrule
    \multirow{2}{*}{SCAN} & Source & \textit{jump twice} \\
    & Target & JUMP JUMP \\
    \midrule
    \multirow{2}{*}{GEO} & Source & \textit{how many people in new york city} \\
    & Target & SELECT CITY alias0 . POPULATION FROM CITY AS CITY  alias0 WHERE CITY alias0 . CITY\_NAME = CITY\_NAME ; \\
    \midrule
    \multirow{2}{*}{ADV} & Source & \textit{Which department includes a history of american film ?} \\
    & Target & SELECT DISTINCT COURSE alias0 . DEPARTMENT FROM COURSE AS COURSE alias0 WHERE COURSE alias0 . NAME LIKE TOPIC ;  \\
    \midrule
    \multirow{2}{*}{Geography} & Source & \textit{how many people live in new york} \\
    & Target & SELECT STATE alias0 . POPULATION FROM STATE AS STATE alias0 WHERE STATE alias0 . STATE\_NAME = " new york " ; \\
    \midrule
    \multirow{4}{*}{Advising} & Source  & \textit{I would like to see A History of American Film courses of 2 credits .} \\
    & \multirow{3}{*}{Target} & SELECT DISTINCT COURSE alias0 . DEPARTMENT , COURSE alias0 . NAME , COURSE alias0 . NUMBER FROM \\
    & & COURSE AS COURSE alias0 WHERE ( COURSE alias0 . DESCRIPTION LIKE "\% A History of American Film \%"  \\
    & & OR COURSE alias0 . NAME LIKE "\% A History of American Film \%" ) AND COURSE alias0 . CREDITS = 2 ; \\
    \bottomrule
  \end{tabular}}
  \vspace{-2mm}
  \caption{Example source and target sequences from SCAN, GEO, ADV, Geography, and Advising.}
  \label{table:data}
\end{table*}

\begin{table*}
  \resizebox{\textwidth}{!}{
  \centering
  \begin{tabular}{lcccll}
    \toprule
    \multirow{2}{*}{\textbf{Data}} & \multirow{2}{*}{\textbf{Primitive}} & \multirow{2}{*}{\textbf{Semantic Links}} & \multirow{2}{*}{\textbf{Variant}} & \multicolumn{2}{c}{\textbf{Concept Rule}} \\
    \cmidrule(r){5-6}
     & & & & \textbf{Primitive Rule} & \textbf{Variant Rule} \\
    \midrule
    \multirow{4}{*}{SCAN} & \textit{jump} & \multirow{4}{*}{Lexical Variant} & \textit{jump\_0} & \textit{jump} $\rightarrow$ JUMP & \textit{jump\_0} $\rightarrow$ JUMP \\
    & \textit{look} & & \textit{look\_0} & \textit{look} $\rightarrow$ LOOK & \textit{look\_0} $\rightarrow$ LOOK \\
    & \textit{run} & & \textit{run\_0} & \textit{run} $\rightarrow$ RUN & \textit{run\_0} $\rightarrow$ RUN \\
    & \textit{walk} & & \textit{walk\_0} & \textit{walk} $\rightarrow$ WALK & \textit{walk\_0} $\rightarrow$ WALK \\
    \midrule
    \multirow{4}{*}{GEO} & \textit{new york city} & \multirow{4}{*}{Co-hyponym} & \textit{houston city} & \textit{new york city} $\rightarrow$ CITY\_NAME & \textit{houston city} $\rightarrow$ CITY\_NAME \\
    & \textit{mississippi rivier} & & \textit{red rivier} & \textit{mississippi rivier} $\rightarrow$ RIVER\_NAME & \textit{red rivier} $\rightarrow$ RIVER\_NAME \\
    & \textit{dc} & & \textit{kansas} & \textit{dc} $\rightarrow$ STATE\_NAME & \textit{kansas} $\rightarrow$ STATE\_NAME \\
    & \textit{dover} & & \textit{salem} & \textit{dover} $\rightarrow$ CAPITAL\_NAME & \textit{salem} $\rightarrow$ CAPITAL\_NAME \\
    \midrule
    \multirow{4}{*}{ADV} & \textit{a history of american film} & \multirow{4}{*}{Co-hyponym} & \textit{advanced ai techniques} & \textit{a history of american film} $\rightarrow$ TOPIC & \textit{advanced ai techniques} $\rightarrow$ TOPIC \\
    & \textit{aaron magid} & & \textit{cargo} & \textit{aaron magid} $\rightarrow$ INSTRUCTOR & \textit{cargo} $\rightarrow$ INSTRUCTOR \\
    & \textit{aaptis} & & \textit{survmeth} & \textit{aaptis} $\rightarrow$ DEPARTMENT & \textit{survmeth} $\rightarrow$ DEPARTMENT \\
    & \textit{100} & & \textit{171} & \textit{100} $\rightarrow$ NUMBER & \textit{171} $\rightarrow$ NUMBER \\
    \bottomrule
  \end{tabular}}
  \vspace{-2mm}
  \caption{Concept rules with primitives and their example variants.}
  \label{table:exp2}
\end{table*}

\textbf{Lexical Variant} refers to an alternative expression form for the same concept, where the various forms may derive from foreign languages, abbreviations, and even mistakes. A basic assumption is that all languages change over time due to non-linguistic factors. Since the rise of sociolinguistics in the 1960s, studies on linguistic variability, a characteristic of language, are central to the language use and motivations for speakers to vary the pronunciation, word choice, or morphology of existing concepts \citep{labov1963social}. Taking ``\textit{United States of America}'' as an example, people have generally accepted the semantic connections among its lexical variants in history, including ``\textit{America}'' and ``\textit{United States}'', as well as the initialisms ``\textit{U.S.}'' and ``\textit{U.S.A}''. Many efforts have been devoted on lexical variants representation \citep{nguyen-grieve-2020-word}, detection \citep{barteld-2017-detecting}, normalization \citep{baldwin-etal-2015-shared} to keep machines up with the trend of the times. \\
\textbf{Co-hyponym} is a linguistic term to designate a semantic relation between two group members belonging to the same broader class, where each member is a hyponym, also called subtype or subordinate, and the class is a hypernym \citep{lyons1995linguistic}. The ``is-a'' hypernymy relation between a generic hypernym and its specific hyponyms builds semantic connections among co-hyponyms. An example of such a hierarchical structure can be ``\textit{Mississippi}'' and ``\textit{California}'' in the domain of ``\textit{state}''. Specifically, ``\textit{Mississippi}'' and ``\textit{California}'' are two hyponyms, and ``\textit{state}'' is a hypernym. Thus, ``\textit{Mississippi}'' and ``\textit{California}'' are semantically connected to be co-hyponyms for each other. Harvesting hypernymy relations \citep{wang2020birre} plays an essential role for downstream knowledge graph construction \citep{9416312}, out-vocabulary generalization \citep{Dash_Chowdhury_Gliozzo_Mihindukulasooriya_Fauceglia_2020}, and taxonomy expansion \citep{yu2020steam}. \\
\textbf{Synonym} stands for a word, morpheme, or phrase that shares exactly or nearly the same semantics with another one. Many tend to assume synonyms are utterances that occur in most contexts in common, so they are semantically closely related enough to be synonyms for each other \citep{10.1145/365628.365657,harris1954distributional}. The existence of the association to contexts is a basic assumption supporting the advance of recent masked language modeling \citep{devlin-etal-2019-bert}. Given that, one of the definitions of a synonymous relation is a semantic link between two expressions if substitution of one for the other never hurts the true value of the context \citep{stanojevic2009cognitive}. For instance, the substitution of ``\textit{heavily populated}'' for ``\textit{populous}'' will seldom alter the truth of the sentence in Figure \ref{figure:ml}. Such semantic similarity can be observed in continuous vector space from a trained representation as well \citep{DBLP:journals/corr/abs-1301-3781}. Synonym discovery \citep{yu2020synet} has been a fundamental job to construct knowledge base and thus benefits substantial researches.

\section{Models} \label{appendix:model}

All models are built within the encoder-decoder framework \citep{sutskever2014sequence}. We reproduce RNN, CNN, and TFM by ourselves to have fewer parameters than the original versions for the experimental purposes. The dropout rate is 0.5 for RNN, CNN, and TFM \citep{srivastava2014dropout}. We implement LSTM, Transformer, and Dynamic Conv. within the library \textit{fairseq}.\footnote{\url{https://github.com/pytorch/fairseq}} \citep{ott2019fairseq} and inherit its default model structures.\footnote{LSTM is adapted from \textit{lstm\_luong\_wmt\_en\_de}; Transformer is adapted from \textit{transformer\_iwslt\_de\_en}; Dynamic Conv. is adapted from \textit{lightconv\_iwslt\_de\_en}.} In contrast to early stopping \citep{prechelt1998early}, we prefer a fixed training regime sufficient enough for models to fully converge in practice with a focus on the systematic generalization observation instead of superior structure exploration. Training is on a single Nvidia Tesla V100. Without specific notes, hyperparameters are shared throughout the work. \\
\textbf{RNN} denotes bi-directional recurrent network \citep{schuster1997bidirectional,hochreiter1997long} with long short-term memory units and an attention mechanism \citep{bahdanau2014neural}. Its encoder consists of two layers with a hidden size of 256 in each direction, and its decoder has one layer with a hidden size of 512. The embedding size is 512 for both encoder and decoder. There are a total of $5.29M$ trainable parameters. Teacher forcing with a rate of $0.5$ serves to spur up the training process \citep{williams1989learning}. \\
\textbf{CNN} denotes the fully convolutional seq2seq network \citep{gehring2017convolutional}. The size of the position embedding layer is 128 for encoding and 256 for decoding, while that of the token embedding layer is 512 for both encoding and decoding. There are 10 convolutional layers with 512 as the hidden size and 3 as the kernel size in both encoder and decoder, resulting in a total of $33.55M$ trainable parameters. \\
\textbf{TFM} denotes transformers, an attention-based network \citep{NIPS2017_7181}. As a tiny version, TFM has 2 layers for each encoder and decoder with 8 attention heads and a dimension of 512. The size of the feedforward layer is 2048. We utilize the cyclic nature of $\sin$ and $\cos$ functions to represent token positions. There are a total of $15.02M$ trainable parameters. \\
\textbf{LSTM} is adapted from the recurrent network used by \citet{luong-etal-2015-effective} for statistical MT. The size of the embedding layer is 1000. There are 4 layers in both encoder and decoder with a hidden size of 512 and a dropout rate of 0.2. \\
\textbf{Transformer}, the same as TFM, is adapted from the base version of transformers in the work of \citet{NIPS2017_7181}, while TFM is a tiny version to test systematic generalization. The dimension is 512 for the embedding layer, 1024 for the feedforward layer, and 512 for the attention layer. There are 6 attention blocks in both encoder and decoder with 4 attention heads and 0.3 dropout probability. \\
\textbf{Dynamic Conv.} is adapted from the seq2seq convolutional network proposed by \citet{wu2018pay}, where the hidden size of the embedding layer, encoder layer, and decoder layer is 512. The number of attention heads is 4, and the dimension of the feedforward layer is 1024 for both encoder and encoder. There are 6 layers in the encoder and 7 layers in the decoder. The dropout rate is 0.1 for both attention and weight units.

\begin{table*}
  \resizebox{\textwidth}{!}{
  \centering
  \begin{tabular}{lcccl}
    \toprule
    \textbf{Data} & \textbf{Primitive} & \textbf{Variant} & \textbf{\#Variants} & \textbf{Prompt}\\
    \midrule
    SCAN & \textit{jump} & \textit{jump\_0} & 10 & \textit{[concept] twice} \\
    \midrule
    \multirow{4}{*}{GEO} & \textit{new york city} & \textit{houston city} & 39 & \textit{how many people in [concept]} \\
     & \textit{mississippi rivier} & \textit{red rivier} & 9 & \textit{how long is [concept]} \\
     & \textit{dc} & \textit{kansas} & 49 & \textit{where is [concept]} \\
     & \textit{dover} & \textit{salem} & 8 & \textit{what states capital is [concept]} \\
    \midrule
    \multirow{4}{*}{ADV} & \textit{a history of american film} & \textit{advanced ai techniques} & 5/424 & \textit{who teaches [concept] ?} \\
    & \textit{aaron magid} & \textit{cargo} & 5/492 & \textit{does [concept] give upper-level courses ?} \\
    & \textit{aaptis} & \textit{survmeth} & 5/1720 & \textit{name core courses for [concept] .} \\
    & \textit{100} & \textit{171} & 5/1895 & \textit{can undergrads take [concept] ?} \\
    \bottomrule
  \end{tabular}}
  \vspace{-2mm}
  \caption{Prompts with example primitives and sampled variants. In SCAN, primitives share the same prompt and the number of variants can be changed. In GEO, we make use of the full variants set. In ADV, we randomly sample 5 variants for each source sequence so that we cover all the variants with a test set of an appropriate size. We generate variant samples by filling the prompt with variants accordingly.}
  \label{table:exp1}
\end{table*}

\begin{table*}[htp!]
  \resizebox{\textwidth}{!}{
  \centering
  \begin{tabular}{llccccccccccccccccccc}
    \toprule
    & & \multicolumn{5}{c}{SCAN} & \multicolumn{5}{c}{GEO} & \multicolumn{5}{c}{ADV} & \multicolumn{2}{c}{Geography} & \multicolumn{2}{c}{Advising} \\
    \cmidrule(r){3-7}
    \cmidrule(r){8-12}
    \cmidrule(r){13-17}
    \cmidrule(r){18-19}
    \cmidrule(r){20-21}
    \multicolumn{2}{l}{\textbf{Data}} & \multicolumn{3}{c}{Exp. 1} & \multicolumn{2}{c}{Exp. 2} & \multicolumn{3}{c}{Exp. 1} & \multicolumn{2}{c}{Exp. 2} & \multicolumn{3}{c}{Exp. 1} & \multicolumn{2}{c}{Exp. 2} & \multirow{2}{*}{Bas.} & \multirow{2}{*}{Aug.} & \multirow{2}{*}{Bas.} & \multirow{2}{*}{Aug.} \\
    \cmidrule(r){3-5}
    \cmidrule(r){6-7}
    \cmidrule(r){8-10}
    \cmidrule(r){11-12}
    \cmidrule(r){13-15}
    \cmidrule(r){16-17}
    & & Sta. & Dif. & Cha. & Sta. & Dif. & Sta. & Dif. & Cha. & Sta. & Dif. & Sta. & Dif. & Cha. & Sta. & Dif. \\
    \midrule
    \multicolumn{2}{l}{\textbf{Train Size}} & 20946 & 20942 & 20928 & 20950 & 20946 & 724 & 720 & 711 & 728 & 724 & 6038 & 6034 & 5969 & 6040 & 6036 & 598 & 701 & 3814 & 5660 \\
    \multicolumn{2}{l}{\textbf{Test Size}}  & 308240 & 308240 & 308240 & 308240 & 308240 & 21350 & 21350 & 21350 & 21350 & 21350 & 107614 & 107614 & 107614 & 107614 & 107614 & 279 & 279 & 573 & 573 \\
    \cmidrule(r){3-7}
    \cmidrule(r){8-12}
    \cmidrule(r){13-17}
    \cmidrule(r){18-18}
    \cmidrule(r){19-19}
    \cmidrule(r){20-20}
    \cmidrule(r){21-21}
    \multirow{3}{*}{\textbf{Time}} & RNN & \multicolumn{5}{c}{21} & \multicolumn{5}{c}{5} & \multicolumn{5}{c}{19} & 4 & 5 & 27 & 35 \\
     & CNN & \multicolumn{5}{c}{17} & \multicolumn{5}{c}{1.2} & \multicolumn{5}{c}{11} & 1 & 1.2 & 12 & 19 \\
     & TFM & \multicolumn{5}{c}{7} & \multicolumn{5}{c}{0.5} & \multicolumn{5}{c}{5} & 0.4 & 0.5 & 6 & 8 \\
    \bottomrule
  \end{tabular}}
  \vspace{-2mm}
  \caption{Data statistics and training time per epoch in seconds. The batch size of each epoch for GEO and Geography is 32, and that for the others is 128.}
  \label{table:details}
\end{table*}

\section{Data} \label{appendix:data}

\textbf{IWSLT} involves IWSLT'14 \citep{cettolo2014report} English-German (En-De) and German-English (De-En), IWSLT'15 \citep{Cettolo2015TheI2} English-French (En-Fr) and French-English (Fr-En) translation tasks. The goal is to translate a sentence from one language to the other. The IWSLT'14 En-De and De-EN have 160,239 sequence pairs for training and 7,283 for validation. We make use of IWSLT14.TED.dev\{2010, 2012\} and IWSLT14.TED.tst\{2010, 2011, 2012\} to measure translation performance, resulting in a total of 6,750 test samples. In terms of IWSLT'15 En-Fr and Fr-En, there are 205,572 sequence pairs for training. We employ IWSLT15.TED.dev2010 and IWSLT15.TED.tst\{2010, 2011, 2012, 2013\} as the validation set and IWSLT15.tst\{2014, 2015\} as the test set. As a consequence, there are 5,519 samples for validation and 2,385 for evaluation. For all four translation tasks, we apply BPE with 10K tokens to share.

\section{Experiments} \label{appendix:exp}

\subsection{Inductive learning}

Semantic linking can be operated via inductive learning, where we replace the concept in the prompt with primitives and their variants. The learning rate to train CNN in GEO is changed to 5$e^{-4}$. Prompts used in SCAN, GEO, and ADV are expressed in Table \ref{table:exp1}. Detailed experimental results with respect to three levels can be found in Table \ref{table:1sta}, Table \ref{table:1dif}, and Table \ref{table:1cha}.

\subsection{Deductive learning}

Semantic linking can be established via deductive learning, where we put concept rules without context information in the training set instead of specific sequence samples. Example concept rules for SCAN, GEO, and ADV are presented in Table \ref{table:exp2}. Detailed experimental results with respect to two levels can be found in Table \ref{table:2sta} and Table \ref{table:2dif}.

\subsection{Sensitivity analysis}

\begin{figure*}[!hbp]
    \begin{center}
        \includegraphics[width=\textwidth]{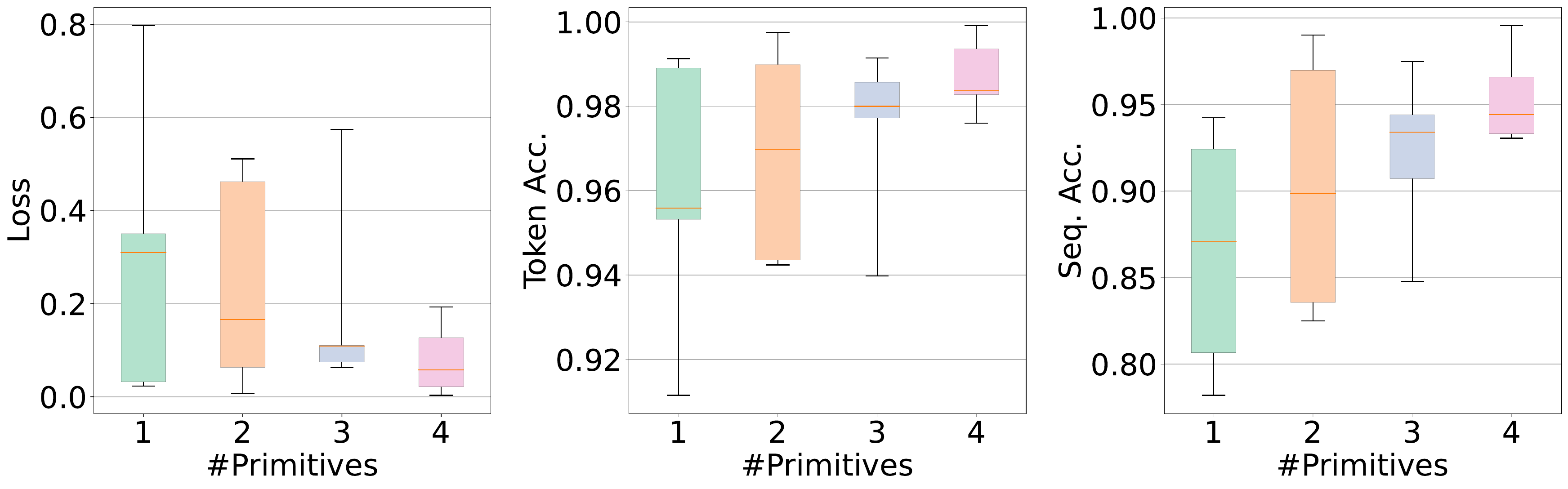}
    \end{center}
    \vspace{-2mm}
    \caption{The complete version of Figure \ref{figure:sa} in Section \ref{sec:sa} regarding \#primitives.}
\label{figure:sa_p}
\vspace{4mm}
    \begin{center}
    \includegraphics[width=\textwidth]{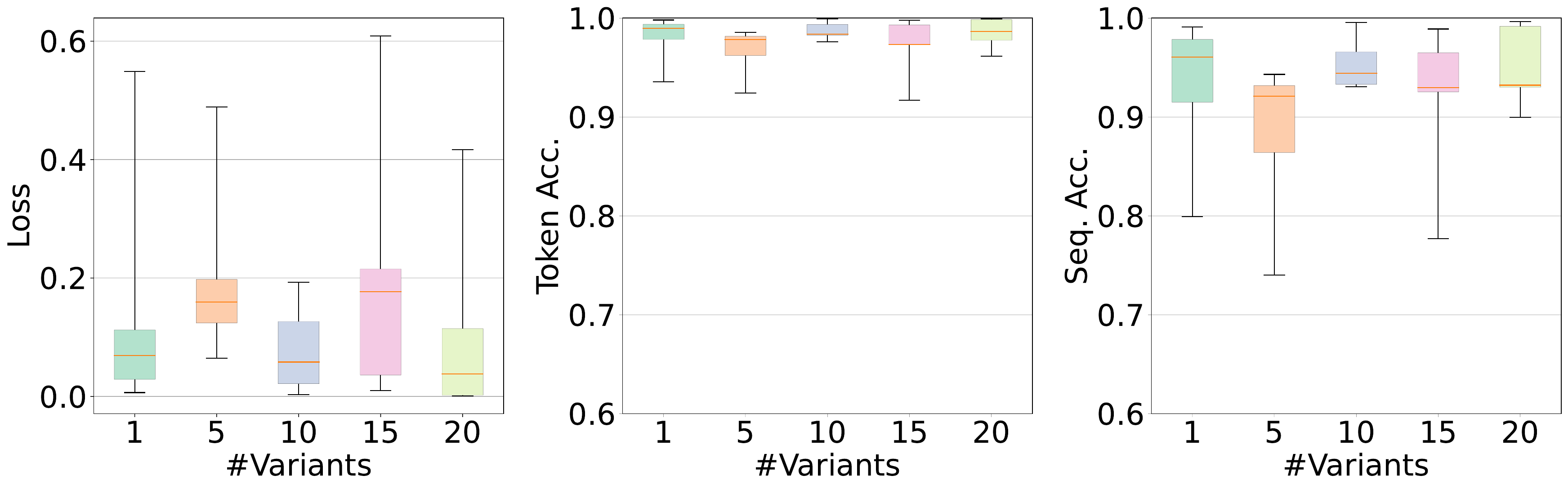}
    \end{center}
    \vspace{-2mm}
    \caption{The complete version of Figure \ref{figure:sa} in Section \ref{sec:sa} regarding \#variants.}
\label{figure:sa_v}
\end{figure*}

In sensitivity analysis, we adjust the number of primitives (\#primitives) and the number of variants per primitive (\#variants) over RNN on SCAN. The complete versions of Figure \ref{figure:sa} in Section \ref{sec:sa} are presented as Figure \ref{figure:sa_p} and Figure \ref{figure:sa_v} for \#primitives and \#variants respectively.

\begin{table*}[!htbp]
  \centering
  \resizebox{0.97\textwidth}{!}{
  \begin{tabular}{llcccccc}
    \toprule
    \multirow{2}{*}{\textbf{Data}} & \multirow{2}{*}{\textbf{Model}} & \multicolumn{3}{c}{\textbf{Train}} & \multicolumn{3}{c}{\textbf{Test}} \\
    \cmidrule(r){3-5}
    \cmidrule(r){6-8}
     & & \textbf{Loss} & \textbf{Token Acc.\%} & \textbf{Seq. Acc.\%} & \textbf{Loss} & \textbf{Token Acc.\%} & \textbf{Seq. Acc.\%}\\
    \midrule
     & RNN & $0.00\pm0.00$ & $100.00\pm0.00$ & $99.99\pm0.02$ & $0.00\pm0.00$ & $99.99\pm0.03$ & $99.95\pm0.08$ \\
    SCAN & CNN & $0.00\pm0.00$ & $99.81\pm0.09$ & $98.78\pm0.55$ & $0.00\pm0.00$ & $99.96\pm0.08$ & $99.85\pm0.34$ \\
     & TFM & $0.00\pm0.00$ & $99.82\pm0.02$ & $98.83\pm0.12$ & $0.06\pm0.03$ & $98.91\pm0.78$ & $97.35\pm1.62$ \\
    \midrule
     & RNN & $0.15\pm0.02$ & $97.73\pm0.42$ & $80.25\pm2.81$ & $1.36\pm0.48$ & $75.71\pm8.42$ & $44.95\pm14.69$ \\
    GEO & CNN & $0.07\pm0.01$ & $98.23\pm0.39$ & $76.80\pm2.25$ & $9.01\pm4.26$ & $87.99\pm2.67$ & $69.46\pm5.78$ \\
     & TFM & $0.02\pm0.00$ & $99.63\pm0.07$ & $91.60\pm1.41$ & $4.55\pm1.39$ & $75.37\pm7.84$ & $45.93\pm12.42$ \\
    \midrule
     & RNN & $0.03\pm0.01$ & $99.40\pm0.13$ & $82.74\pm2.78$ & $6.04\pm0.95$ & $58.61\pm6.18$ & $36.18\pm5.75$ \\
    ADV & CNN & $0.01\pm0.01$ & $99.59\pm0.07$ & $85.13\pm1.95$ & $23.56\pm4.95$ & $57.83\pm7.55$ & $45.08\pm9.32$ \\
     & TFM & $0.00\pm0.00$ & $99.92\pm0.01$ & $96.14\pm0.28$ & $15.12\pm1.00$ & $53.43\pm2.80$ & $42.59\pm3.65$ \\
    \bottomrule
  \end{tabular}}
  \vspace{-2mm}
  \caption{Results of Standard inductive learning.}
  \label{table:1sta}
\vspace{2mm}
  \centering
  \resizebox{0.97\textwidth}{!}{
  \begin{tabular}{llcccccc}
    \toprule
    \multirow{2}{*}{\textbf{Data}} & \multirow{2}{*}{\textbf{Model}} & \multicolumn{3}{c}{\textbf{Train}} & \multicolumn{3}{c}{\textbf{Test}} \\
    \cmidrule(r){3-5}
    \cmidrule(r){6-8}
     & & \textbf{Loss} & \textbf{Token Acc.\%} & \textbf{Seq. Acc.\%} & \textbf{Loss} & \textbf{Token Acc.\%} & \textbf{Seq. Acc.\%}\\
    \midrule
     & RNN & $0.00\pm0.00$ & $100.00\pm0.00$ & $99.99\pm0.01$ & $0.00\pm0.00$ & $99.96\pm0.02$ & $99.85\pm0.08$ \\
    SCAN & CNN & $0.00\pm0.00$ & $99.77\pm0.19$ & $98.62\pm1.13$ & $0.03\pm0.06$ & $99.76\pm0.54$ & $99.52\pm1.07$ \\
     & TFM & $0.00\pm0.00$ & $99.79\pm0.03$ & $98.59\pm0.12$ & $0.06\pm0.03$ & $98.90\pm1.10$ & $96.86\pm2.64$ \\
    \midrule
     & RNN & $0.16\pm0.03$ & $97.39\pm0.67$ & $78.33\pm4.31$ & $1.29\pm0.27$ & $75.69\pm6.12$ & $43.27\pm13.47$ \\
    GEO & CNN & $0.07\pm0.01$ & $98.25\pm0.13$ & $76.53\pm1.68$ & $13.87\pm3.19$ & $79.51\pm6.03$ & $51.20\pm8.64$\\
     & TFM & $0.00\pm0.11$ & $99.60\pm0.11$ & $91.33\pm1.46$ & $4.50\pm0.80$ & $75.11\pm4.88$ & $44.59\pm9.76$ \\
    \midrule
     & RNN & $0.03\pm0.01$ & $99.26\pm0.21$ & $79.57\pm4.12$ & $5.80\pm0.92$ & $59.74\pm5.67$ & $35.69\pm6.05$ \\
    ADV & CNN & $0.02\pm0.00$ & $99.56\pm0.05$ & $84.06\pm1.57$ & $24.58\pm3.40$ & $54.05\pm5.74$ & $42.14\pm6.90$ \\
     & TFM & $0.00\pm0.00$ & $99.91\pm0.01$ & $95.88\pm0.23$ & $15.84\pm1.51$ & $51.51\pm4.50$ & $41.28\pm4.35$ \\
    \bottomrule
  \end{tabular}}
  \vspace{-2mm}
  \caption{Results of Difficult inductive learning.}
  \label{table:1dif}
\vspace{2mm}
  \centering
  \resizebox{0.97\textwidth}{!}{
  \begin{tabular}{llcccccc}
    \toprule
    \multirow{2}{*}{\textbf{Data}} & \multirow{2}{*}{\textbf{Model}} & \multicolumn{3}{c}{\textbf{Train}} & \multicolumn{3}{c}{\textbf{Test}} \\
    \cmidrule(r){3-5}
    \cmidrule(r){6-8}
     & & \textbf{Loss} & \textbf{Token Acc.\%} & \textbf{Seq. Acc.\%} & \textbf{Loss} & \textbf{Token Acc.\%} & \textbf{Seq. Acc.\%}\\
    \midrule
     & RNN & $0.00\pm0.00$ & $100.00\pm0.00$ & $99.99\pm0.02$ & $0.20\pm0.45$ & $99.95\pm0.08$ & $99.80\pm0.31$ \\
    SCAN & CNN & $0.00\pm0.00$ & $99.85\pm0.05$ & $99.00\pm0.30$ & $0.14\pm0.31$ & $98.89\pm2.44$ & $97.57\pm5.24$ \\
     & TFM & $0.00\pm0.00$ & $99.82\pm0.05$ & $98.85\pm0.27$ & $0.07\pm0.05$ & $98.76\pm0.85$ & $96.38\pm2.81$ \\
    \midrule
     & RNN & $0.15\pm0.04$ & $97.76\pm0.74$ & $79.77\pm4.19$ & $1.52\pm0.29$ & $73.46\pm3.05$ & $36.77\pm5.60$ \\
    GEO & CNN & $0.07\pm0.01$ & $98.23\pm0.17$ & $75.98\pm1.46$ & $15.83\pm4.56$ & $77.40\pm2.48$ & $48.53\pm3.40$ \\
     & TFM & $0.02\pm0.00$ & $99.60\pm0.06$ & $91.00\pm1.20$ & $6.01\pm1.03$ & $68.41\pm4.76$ & $36.93\pm7.47$ \\
    \midrule
     & RNN & $0.03\pm0.01$ & $99.23\pm0.13$ & $79.90\pm1.85$ & $5.95\pm0.90$ & $58.11\pm5.82$ & $35.45\pm6.69$ \\
    ADV & CNN & $0.01\pm0.01$ & $99.68\pm0.15$ & $87.90\pm5.05$ & $23.08\pm6.34$ & $53.66\pm2.57$ & $41.37\pm4.04$ \\
     & TFM & $0.00\pm0.00$ & $99.93\pm0.01$ & $96.41\pm0.24$ & $16.59\pm0.98$ & $49.17\pm2.58$ & $38.88\pm2.68$ \\
    \bottomrule
  \end{tabular}}
  \vspace{-2mm}
  \caption{Results of Challenging inductive learning.}
  \label{table:1cha}
\vspace{2mm}
  \centering
  \resizebox{0.97\textwidth}{!}{
  \begin{tabular}{llcccccc}
    \toprule
    \multirow{2}{*}{\textbf{Data}} & \multirow{2}{*}{\textbf{Model}} & \multicolumn{3}{c}{\textbf{Train}} & \multicolumn{3}{c}{\textbf{Test}} \\
    \cmidrule(r){3-5}
    \cmidrule(r){6-8}
     & & \textbf{Loss} & \textbf{Token Acc.\%} & \textbf{Seq. Acc.\%} & \textbf{Loss} & \textbf{Token Acc.\%} & \textbf{Seq. Acc.\%}\\
    \midrule
     & RNN & $0.00\pm0.00$ & $99.99\pm0.03$ & $99.90\pm0.23$ & $0.05\pm0.06$ & $99.48\pm0.71$ & $98.27\pm2.38$ \\
    SCAN & CNN & $0.00\pm0.00$ & $99.79\pm0.14$ & $98.78\pm0.79$ & $0.00\pm0.00$ & $99.99\pm0.01$ & $99.96\pm0.03$ \\
     & TFM & $0.00\pm0.00$ & $99.82\pm0.03$ & $98.78\pm0.17$ & $0.27\pm0.22$ & $96.90\pm1.78$ & $91.94\pm4.04$ \\
    \midrule
     & RNN & $0.17\pm0.03$ & $97.50\pm0.30$ & $78.54\pm2.16$ & $2.83\pm0.69$ & $54.44\pm7.15$ & $13.61\pm7.08$ \\
    GEO & CNN & $0.08\pm0.01$ & $97.97\pm0.24$ & $77.03\pm1.42$ & $51.08\pm25.97$ & $41.86\pm3.38$ & $4.85\pm4.66$ \\
     & TFM & $0.02\pm0.00$ & $99.54\pm0.31$ & $91.82\pm2.27$ & $6.03\pm1.56$ & $67.02\pm6.91$ & $36.38\pm10.08$ \\
    \midrule
     & RNN & $0.08\pm0.02$ & $98.64\pm0.31$ & $68.84\pm4.57$ & $7.95\pm1.13$ & $36.50\pm7.66$ & $12.84\pm4.31$ \\
    ADV & CNN & $0.02\pm0.00$ & $99.53\pm0.07$ & $84.64\pm1.20$ & $31.12\pm4.76$ & $43.51\pm11.31$ & $32.33\pm12.93$ \\
     & TFM & $0.00\pm0.00$ & $99.91\pm0.02$ & $96.33\pm0.37$ & $13.72\pm1.41$ & $56.82\pm3.79$ & $47.43\pm3.71$ \\
    \bottomrule
  \end{tabular}}
  \vspace{-2mm}
  \caption{Results of Standard deductive learning.}
  \label{table:2sta}
\end{table*}

\begin{table*}[!htbp]
  \resizebox{\textwidth}{!}{
  \centering
  \begin{tabular}{llcccccc}
    \toprule
    \multirow{2}{*}{\textbf{Data}} & \multirow{2}{*}{\textbf{Model}} & \multicolumn{3}{c}{\textbf{Train}} & \multicolumn{3}{c}{\textbf{Test}} \\
    \cmidrule(r){3-5}
    \cmidrule(r){6-8}
     & & \textbf{Loss} & \textbf{Token Acc.\%} & \textbf{Seq. Acc.\%} & \textbf{Loss} & \textbf{Token Acc.\%} & \textbf{Seq. Acc.\%}\\
    \midrule
     & RNN & $0.00\pm0.00$ & $99.99\pm0.01$ & $99.95\pm0.07$ & $0.08\pm0.08$ & $98.70\pm0.92$ & $95.39\pm2.72$ \\
    SCAN & CNN & $0.00\pm0.00$ & $99.62\pm0.34$ & $98.82\pm1.09$ & $0.13\pm0.29$ & $98.59\pm3.10$ & $96.66\pm7.27$ \\
     & TFM & $0.00\pm0.00$ & $99.82\pm0.03$ & $98.78\pm0.12$ & $0.21\pm0.20$ & $96.68\pm2.21$ & $91.26\pm5.80$ \\
    \midrule
     & RNN & $0.20\pm0.03$ & $96.93\pm0.71$ & $75.35\pm3.57$ & $4.40\pm2.50$ & $39.71\pm18.38$ & $7.67\pm5.34$ \\
    GEO & CNN & $0.08\pm0.01$ & $97.77\pm0.76$ & $76.41\pm2.80$ & $32.94\pm4.26$ & $41.07\pm7.48$ & $4.04\pm2.18$ \\
     & TFM & $0.02\pm0.00$ & $99.56\pm0.11$ & $91.08\pm1.56$ & $5.97\pm1.05$ & $65.97\pm5.17$ & $31.57\pm7.42$ \\
    \midrule
     & RNN & $0.08\pm0.02$ & $98.54\pm0.28$ & $67.10\pm3.45$ & $7.87\pm1.01$ & $36.42\pm7.39$ & $12.66\pm5.19$ \\
    ADV & CNN & $0.04\pm0.05$ & $98.78\pm1.91$ & $77.14\pm23.28$ & $32.44\pm6.07$ & $35.34\pm14.68$ & $23.58\pm16.04$ \\
     & TFM & $0.00\pm0.00$ & $99.92\pm0.02$ & $96.41\pm0.26$ & $14.92\pm1.31$ & $53.33\pm3.85$ & $43.24\pm5.14$ \\
    \bottomrule
  \end{tabular}}
  \vspace{-2mm}
  \caption{Results of Difficult deductive learning.}
  \label{table:2dif}
\end{table*}

%% file: 0.main.bbl
\begin{thebibliography}{84}
\expandafter\ifx\csname natexlab\endcsname\relax\def\natexlab#1{#1}\fi

\bibitem[{Aky{\"u}rek et~al.(2021)Aky{\"u}rek, Aky{\"u}rek, and
  Andreas}]{rek2021learning}
Ekin Aky{\"u}rek, Afra~Feyza Aky{\"u}rek, and Jacob Andreas. 2021.
\newblock Learning to recombine and resample data for compositional
  generalization.
\newblock In \emph{International Conference on Learning Representations}.

\bibitem[{Akyurek and Andreas(2021)}]{akyurek-andreas-2021-lexicon}
Ekin Akyurek and Jacob Andreas. 2021.
\newblock \href {https://doi.org/10.18653/v1/2021.acl-long.382} {Lexicon
  learning for few shot sequence modeling}.
\newblock In \emph{Proceedings of the 59th Annual Meeting of the Association
  for Computational Linguistics and the 11th International Joint Conference on
  Natural Language Processing (Volume 1: Long Papers)}, pages 4934--4946,
  Online. Association for Computational Linguistics.

\bibitem[{Andreas(2020)}]{andreas-2020-good}
Jacob Andreas. 2020.
\newblock \href {https://doi.org/10.18653/v1/2020.acl-main.676} {Good-enough
  compositional data augmentation}.
\newblock In \emph{Proceedings of the 58th Annual Meeting of the Association
  for Computational Linguistics}, pages 7556--7566, Online. Association for
  Computational Linguistics.

\bibitem[{Arthur et~al.(2016)Arthur, Neubig, and
  Nakamura}]{arthur-etal-2016-incorporating}
Philip Arthur, Graham Neubig, and Satoshi Nakamura. 2016.
\newblock \href {https://doi.org/10.18653/v1/D16-1162} {Incorporating discrete
  translation lexicons into neural machine translation}.
\newblock In \emph{Proceedings of the 2016 Conference on Empirical Methods in
  Natural Language Processing}, pages 1557--1567, Austin, Texas. Association
  for Computational Linguistics.

\bibitem[{Auersperger and Pecina(2021)}]{auersperger-pecina-2021-solving}
Michal Auersperger and Pavel Pecina. 2021.
\newblock \href {https://aclanthology.org/2021.ranlp-1.11} {Solving {SCAN}
  tasks with data augmentation and input embeddings}.
\newblock In \emph{Proceedings of the International Conference on Recent
  Advances in Natural Language Processing (RANLP 2021)}, pages 86--91, Held
  Online. INCOMA Ltd.

\bibitem[{Ausubel(1963)}]{ausubel1963psychology}
D.P. Ausubel. 1963.
\newblock \href {https://books.google.ca/books?id=ydzRAAAAMAAJ} {\emph{The
  Psychology of Meaningful Verbal Learning}}.
\newblock Grune \& Stratton.

\bibitem[{Bahdanau et~al.(2015)Bahdanau, Cho, and Bengio}]{bahdanau2014neural}
Dzmitry Bahdanau, Kyunghyun Cho, and Yoshua Bengio. 2015.
\newblock Neural machine translation by jointly learning to align and
  translate.
\newblock In \emph{3rd International Conference on Learning Representations,
  {ICLR} 2015, San Diego, CA, USA, May 7-9, 2015, Conference Track
  Proceedings}.

\bibitem[{Baldwin et~al.(2015)Baldwin, de~Marneffe, Han, Kim, Ritter, and
  Xu}]{baldwin-etal-2015-shared}
Timothy Baldwin, Marie~Catherine de~Marneffe, Bo~Han, Young-Bum Kim, Alan
  Ritter, and Wei Xu. 2015.
\newblock \href {https://doi.org/10.18653/v1/W15-4319} {Shared tasks of the
  2015 workshop on noisy user-generated text: {T}witter lexical normalization
  and named entity recognition}.
\newblock In \emph{Proceedings of the Workshop on Noisy User-generated Text},
  pages 126--135, Beijing, China. Association for Computational Linguistics.

\bibitem[{Barteld(2017)}]{barteld-2017-detecting}
Fabian Barteld. 2017.
\newblock Detecting spelling variants in non-standard texts.
\newblock In \emph{Proceedings of the Student Research Workshop at the 15th
  Conference of the {E}uropean Chapter of the Association for Computational
  Linguistics}, pages 11--22, Valencia, Spain. Association for Computational
  Linguistics.

\bibitem[{Bastings et~al.(2018)Bastings, Baroni, Weston, Cho, and
  Kiela}]{bastings-etal-2018-jump}
Joost Bastings, Marco Baroni, Jason Weston, Kyunghyun Cho, and Douwe Kiela.
  2018.
\newblock Jump to better conclusions: {SCAN} both left and right.
\newblock In \emph{Proceedings of the 2018 {EMNLP} Workshop {B}lackbox{NLP}:
  Analyzing and Interpreting Neural Networks for {NLP}}, pages 47--55,
  Brussels, Belgium. Association for Computational Linguistics.

\bibitem[{Brakel and Frank(2009)}]{brakel2009strong}
Phil{\'e}mon Brakel and Stefan Frank. 2009.
\newblock Strong systematicity in sentence processing by simple recurrent
  networks.
\newblock In \emph{31th Annual Conference of the Cognitive Science Society
  (COGSCI-2009)}, pages 1599--1604. Cognitive Science Society.

\bibitem[{Cettolo et~al.(2015)Cettolo, Niehues, St{\"u}ker, Bentivogli,
  Cattoni, and Federico}]{Cettolo2015TheI2}
M.~Cettolo, J.~Niehues, S.~St{\"u}ker, L.~Bentivogli, R.~Cattoni, and Marcello
  Federico. 2015.
\newblock The iwslt 2015 evaluation campaign.
\newblock In \emph{IWSLT}.

\bibitem[{Cettolo et~al.(2014)Cettolo, Niehues, St{\"u}ker, Bentivogli, and
  Federico}]{cettolo2014report}
Mauro Cettolo, Jan Niehues, Sebastian St{\"u}ker, Luisa Bentivogli, and
  Marcello Federico. 2014.
\newblock Report on the 11th iwslt evaluation campaign, iwslt 2014.
\newblock In \emph{Proceedings of the International Workshop on Spoken Language
  Translation, Hanoi, Vietnam}, volume~57.

\bibitem[{Chaabouni et~al.(2021)Chaabouni, Dess{\`\i}, and
  Kharitonov}]{chaabouni-etal-2021-transformers}
Rahma Chaabouni, Roberto Dess{\`\i}, and Eugene Kharitonov. 2021.
\newblock \href {https://doi.org/10.18653/v1/2021.blackboxnlp-1.9} {Can
  transformers jump around right in natural language? assessing performance
  transfer from {SCAN}}.
\newblock In \emph{Proceedings of the Fourth BlackboxNLP Workshop on Analyzing
  and Interpreting Neural Networks for NLP}, pages 136--148, Punta Cana,
  Dominican Republic. Association for Computational Linguistics.

\bibitem[{Chen et~al.(2020)Chen, Liang, Yu, Song, and Zhou}]{ChenLYSZ20}
Xinyun Chen, Chen Liang, Adams~Wei Yu, Dawn Song, and Denny Zhou. 2020.
\newblock Compositional generalization via neural-symbolic stack machines.
\newblock In \emph{Advances in Neural Information Processing Systems 33: Annual
  Conference on Neural Information Processing Systems 2020, NeurIPS 2020,
  December 6-12, 2020, virtual}.

\bibitem[{Chomsky(1956)}]{chomsky1957syntactic}
Noam Chomsky. 1956.
\newblock Syntactic structures.

\bibitem[{Conklin et~al.(2021)Conklin, Wang, Smith, and
  Titov}]{conklin-etal-2021-meta}
Henry Conklin, Bailin Wang, Kenny Smith, and Ivan Titov. 2021.
\newblock \href {https://doi.org/10.18653/v1/2021.acl-long.258} {Meta-learning
  to compositionally generalize}.
\newblock In \emph{Proceedings of the 59th Annual Meeting of the Association
  for Computational Linguistics and the 11th International Joint Conference on
  Natural Language Processing (Volume 1: Long Papers)}, pages 3322--3335,
  Online. Association for Computational Linguistics.

\bibitem[{Csord\'as et~al.(2021)Csord\'as, Irie, and
  Schmidhuber}]{csordas2021devil}
R\'obert Csord\'as, Kazuki Irie, and J\"urgen Schmidhuber. 2021.
\newblock The devil is in the detail: Simple tricks improve systematic
  generalization of transformers.
\newblock In \emph{Proc. Conf. on Empirical Methods in Natural Language
  Processing (EMNLP)}, Punta Cana, Dominican Republic.

\bibitem[{Dankers et~al.(2022{\natexlab{a}})Dankers, Bruni, and
  Hupkes}]{dankers-etal-2022-paradox}
Verna Dankers, Elia Bruni, and Dieuwke Hupkes. 2022{\natexlab{a}}.
\newblock \href {https://doi.org/10.18653/v1/2022.acl-long.286} {The paradox of
  the compositionality of natural language: A neural machine translation case
  study}.
\newblock In \emph{Proceedings of the 60th Annual Meeting of the Association
  for Computational Linguistics (Volume 1: Long Papers)}, pages 4154--4175,
  Dublin, Ireland. Association for Computational Linguistics.

\bibitem[{Dankers et~al.(2022{\natexlab{b}})Dankers, Lucas, and
  Titov}]{dankers-etal-2022-transformer}
Verna Dankers, Christopher Lucas, and Ivan Titov. 2022{\natexlab{b}}.
\newblock \href {https://doi.org/10.18653/v1/2022.acl-long.252} {Can
  transformer be too compositional? analysing idiom processing in neural
  machine translation}.
\newblock In \emph{Proceedings of the 60th Annual Meeting of the Association
  for Computational Linguistics (Volume 1: Long Papers)}, pages 3608--3626,
  Dublin, Ireland. Association for Computational Linguistics.

\bibitem[{Dash et~al.(2020)Dash, Chowdhury, Gliozzo, Mihindukulasooriya, and
  Fauceglia}]{Dash_Chowdhury_Gliozzo_Mihindukulasooriya_Fauceglia_2020}
Sarthak Dash, Md~Faisal~Mahbub Chowdhury, Alfio Gliozzo, Nandana
  Mihindukulasooriya, and Nicolas~Rodolfo Fauceglia. 2020.
\newblock \href {https://doi.org/10.1609/aaai.v34i05.6263} {Hypernym detection
  using strict partial order networks}.
\newblock \emph{Proceedings of the AAAI Conference on Artificial Intelligence},
  34(05):7626--7633.

\bibitem[{Devlin et~al.(2019)Devlin, Chang, Lee, and
  Toutanova}]{devlin-etal-2019-bert}
Jacob Devlin, Ming-Wei Chang, Kenton Lee, and Kristina Toutanova. 2019.
\newblock \href {https://doi.org/10.18653/v1/N19-1423} {{BERT}: Pre-training of
  deep bidirectional transformers for language understanding}.
\newblock In \emph{Proceedings of the 2019 Conference of the North {A}merican
  Chapter of the Association for Computational Linguistics: Human Language
  Technologies, Volume 1 (Long and Short Papers)}, pages 4171--4186,
  Minneapolis, Minnesota. Association for Computational Linguistics.

\bibitem[{Finegan-Dollak et~al.(2018)Finegan-Dollak, Kummerfeld, Zhang,
  Ramanathan, Sadasivam, Zhang, and Radev}]{data-sql-advising}
Catherine Finegan-Dollak, Jonathan~K. Kummerfeld, Li~Zhang, Karthik Ramanathan,
  Sesh Sadasivam, Rui Zhang, and Dragomir Radev. 2018.
\newblock \href {https://doi.org/10.18653/v1/P18-1033} {Improving text-to-{SQL}
  evaluation methodology}.
\newblock In \emph{Proceedings of the 56th Annual Meeting of the Association
  for Computational Linguistics (Volume 1: Long Papers)}, pages 351--360,
  Melbourne, Australia. Association for Computational Linguistics.

\bibitem[{Fodor and Lepore(2002)}]{fodor2002compositionality}
Jerry~A Fodor and Ernest Lepore. 2002.
\newblock \emph{The compositionality papers}.
\newblock Oxford University Press.

\bibitem[{Fodor and Pylyshyn(1988)}]{fodor1988connectionism}
Jerry~A Fodor and Zenon~W Pylyshyn. 1988.
\newblock Connectionism and cognitive architecture: A critical analysis.
\newblock \emph{Cognition}, 28(1-2):3--71.

\bibitem[{Frank et~al.(2009)Frank, Haselager, and van
  Rooij}]{frank2009connectionist}
Stefan~L Frank, Willem~FG Haselager, and Iris van Rooij. 2009.
\newblock Connectionist semantic systematicity.
\newblock \emph{Cognition}, 110(3):358--379.

\bibitem[{Gehring et~al.(2017)Gehring, Auli, Grangier, Yarats, and
  Dauphin}]{gehring2017convolutional}
Jonas Gehring, Michael Auli, David Grangier, Denis Yarats, and Yann~N Dauphin.
  2017.
\newblock Convolutional sequence to sequence learning.
\newblock In \emph{International Conference on Machine Learning}, pages
  1243--1252. PMLR.

\bibitem[{Gordon et~al.(2020)Gordon, Lopez-Paz, Baroni, and
  Bouchacourt}]{Gordon2020Permutation}
Jonathan Gordon, David Lopez-Paz, Marco Baroni, and Diane Bouchacourt. 2020.
\newblock \href {https://openreview.net/forum?id=SylVNerFvr} {Permutation
  equivariant models for compositional generalization in language}.
\newblock In \emph{International Conference on Learning Representations}.

\bibitem[{Hadley(1994)}]{hadley1994systematicity}
Robert~F Hadley. 1994.
\newblock Systematicity in connectionist language learning.
\newblock \emph{Mind \& Language}, 9(3):247--272.

\bibitem[{Hammerly(1975)}]{hammerly1975deduction}
Hector Hammerly. 1975.
\newblock The deduction/induction controversy.
\newblock \emph{The Modern Language Journal}, 59(1/2):15--18.

\bibitem[{Harris(1954)}]{harris1954distributional}
Zellig~S Harris. 1954.
\newblock Distributional structure.
\newblock \emph{Word}, 10(2-3):146--162.

\bibitem[{Hochreiter and Schmidhuber(1997)}]{hochreiter1997long}
Sepp Hochreiter and J{\"u}rgen Schmidhuber. 1997.
\newblock Long short-term memory.
\newblock \emph{Neural computation}, 9(8):1735--1780.

\bibitem[{Hupkes et~al.(2020)Hupkes, Dankers, Mul, and
  Bruni}]{hupkes2020compositionality}
Dieuwke Hupkes, Verna Dankers, Mathijs Mul, and Elia Bruni. 2020.
\newblock Compositionality decomposed: how do neural networks generalise?
\newblock \emph{Journal of Artificial Intelligence Research}, 67:757--795.

\bibitem[{Ji et~al.(2021)Ji, Pan, Cambria, Marttinen, and Yu}]{9416312}
Shaoxiong Ji, Shirui Pan, Erik Cambria, Pekka Marttinen, and Philip~S. Yu.
  2021.
\newblock \href {https://doi.org/10.1109/TNNLS.2021.3070843} {A survey on
  knowledge graphs: Representation, acquisition, and applications}.
\newblock \emph{IEEE Transactions on Neural Networks and Learning Systems},
  pages 1--21.

\bibitem[{Jiang and Bansal(2021)}]{jiang-bansal-2021-inducing}
Yichen Jiang and Mohit Bansal. 2021.
\newblock \href {https://doi.org/10.18653/v1/2021.emnlp-main.505} {Inducing
  transformer{'}s compositional generalization ability via auxiliary sequence
  prediction tasks}.
\newblock In \emph{Proceedings of the 2021 Conference on Empirical Methods in
  Natural Language Processing}, pages 6253--6265, Online and Punta Cana,
  Dominican Republic. Association for Computational Linguistics.

\bibitem[{Keysers et~al.(2020)Keysers, Sch{\"a}rli, Scales, Buisman, Furrer,
  Kashubin, Momchev, Sinopalnikov, Stafiniak, Tihon, Tsarkov, Wang, van Zee,
  and Bousquet}]{keysers2020measuring}
Daniel Keysers, Nathanael Sch{\"a}rli, Nathan Scales, Hylke Buisman, Daniel
  Furrer, Sergii Kashubin, Nikola Momchev, Danila Sinopalnikov, Lukasz
  Stafiniak, Tibor Tihon, Dmitry Tsarkov, Xiao Wang, Marc van Zee, and Olivier
  Bousquet. 2020.
\newblock Measuring compositional generalization: A comprehensive method on
  realistic data.
\newblock In \emph{International Conference on Learning Representations}.

\bibitem[{Kim and Linzen(2020)}]{kim-linzen-2020-cogs}
Najoung Kim and Tal Linzen. 2020.
\newblock \href {https://doi.org/10.18653/v1/2020.emnlp-main.731} {{COGS}: A
  compositional generalization challenge based on semantic interpretation}.
\newblock In \emph{Proceedings of the 2020 Conference on Empirical Methods in
  Natural Language Processing (EMNLP)}, pages 9087--9105, Online. Association
  for Computational Linguistics.

\bibitem[{Kim(2021)}]{kim2021sequence}
Yoon Kim. 2021.
\newblock Sequence-to-sequence learning with latent neural grammars.
\newblock \emph{Advances in Neural Information Processing Systems}, 34.

\bibitem[{Kingma and Ba(2015)}]{DBLP:journals/corr/KingmaB14}
Diederik~P. Kingma and Jimmy Ba. 2015.
\newblock Adam: {A} method for stochastic optimization.
\newblock In \emph{3rd International Conference on Learning Representations,
  {ICLR} 2015, San Diego, CA, USA, May 7-9, 2015, Conference Track
  Proceedings}.

\bibitem[{Labov(1963)}]{labov1963social}
William Labov. 1963.
\newblock The social motivation of a sound change.
\newblock \emph{Word}, 19(3):273--309.

\bibitem[{Lake(2019)}]{NEURIPS2019_f4d0e2e7}
Brenden~M Lake. 2019.
\newblock Compositional generalization through meta sequence-to-sequence
  learning.
\newblock In \emph{Advances in Neural Information Processing Systems},
  volume~32. Curran Associates, Inc.

\bibitem[{Lake and Baroni(2018)}]{lake2017generalization}
Brenden~M. Lake and Marco Baroni. 2018.
\newblock \href {http://proceedings.mlr.press/v80/lake18a.html} {Generalization
  without systematicity: On the compositional skills of sequence-to-sequence
  recurrent networks}.
\newblock In \emph{Proceedings of the 35th International Conference on Machine
  Learning, {ICML} 2018, Stockholmsm{\"{a}}ssan, Stockholm, Sweden, July 10-15,
  2018}, volume~80 of \emph{Proceedings of Machine Learning Research}, pages
  2879--2888. {PMLR}.

\bibitem[{Li et~al.(2021)Li, Yin, Chen, and Zhang}]{li-etal-2021-compositional}
Yafu Li, Yongjing Yin, Yulong Chen, and Yue Zhang. 2021.
\newblock \href {https://doi.org/10.18653/v1/2021.acl-long.368} {On
  compositional generalization of neural machine translation}.
\newblock In \emph{Proceedings of the 59th Annual Meeting of the Association
  for Computational Linguistics and the 11th International Joint Conference on
  Natural Language Processing (Volume 1: Long Papers)}, pages 4767--4780,
  Online. Association for Computational Linguistics.

\bibitem[{Li et~al.(2019)Li, Zhao, Wang, and
  Hestness}]{li-etal-2019-compositional}
Yuanpeng Li, Liang Zhao, Jianyu Wang, and Joel Hestness. 2019.
\newblock \href {https://doi.org/10.18653/v1/D19-1438} {Compositional
  generalization for primitive substitutions}.
\newblock In \emph{Proceedings of the 2019 Conference on Empirical Methods in
  Natural Language Processing and the 9th International Joint Conference on
  Natural Language Processing (EMNLP-IJCNLP)}, pages 4293--4302, Hong Kong,
  China. Association for Computational Linguistics.

\bibitem[{Liu et~al.(2020)Liu, An, Lou, Chen, Lin, Gao, Zhou, Zheng, and
  Zhang}]{NEURIPS2020_83adc922}
Qian Liu, Shengnan An, Jian-Guang Lou, Bei Chen, Zeqi Lin, Yan Gao, Bin Zhou,
  Nanning Zheng, and Dongmei Zhang. 2020.
\newblock \href
  {https://proceedings.neurips.cc/paper/2020/file/83adc9225e4deb67d7ce42d58fe5157c-Paper.pdf}
  {Compositional generalization by learning analytical expressions}.
\newblock In \emph{Advances in Neural Information Processing Systems},
  volume~33, pages 11416--11427. Curran Associates, Inc.

\bibitem[{Loula et~al.(2018)Loula, Baroni, and
  Lake}]{loula-etal-2018-rearranging}
Joao Loula, Marco Baroni, and Brenden Lake. 2018.
\newblock Rearranging the familiar: Testing compositional generalization in
  recurrent networks.
\newblock In \emph{Proceedings of the 2018 {EMNLP} Workshop {B}lackbox{NLP}:
  Analyzing and Interpreting Neural Networks for {NLP}}, pages 108--114,
  Brussels, Belgium. Association for Computational Linguistics.

\bibitem[{Luong et~al.(2015)Luong, Pham, and
  Manning}]{luong-etal-2015-effective}
Thang Luong, Hieu Pham, and Christopher~D. Manning. 2015.
\newblock \href {https://doi.org/10.18653/v1/D15-1166} {Effective approaches to
  attention-based neural machine translation}.
\newblock In \emph{Proceedings of the 2015 Conference on Empirical Methods in
  Natural Language Processing}, pages 1412--1421, Lisbon, Portugal. Association
  for Computational Linguistics.

\bibitem[{Lyons and John(1995)}]{lyons1995linguistic}
John Lyons and Lyons John. 1995.
\newblock \emph{Linguistic semantics: An introduction}.
\newblock Cambridge University Press.

\bibitem[{Marcus(1998)}]{marcus1998rethinking}
Gary~F Marcus. 1998.
\newblock Rethinking eliminative connectionism.
\newblock \emph{Cognitive psychology}, 37(3):243--282.

\bibitem[{Marcus(2018)}]{marcus2018algebraic}
Gary~F Marcus. 2018.
\newblock \emph{The algebraic mind: Integrating connectionism and cognitive
  science}.
\newblock MIT press.

\bibitem[{Mayer(2002)}]{mayer2002rote}
Richard~E Mayer. 2002.
\newblock Rote versus meaningful learning.
\newblock \emph{Theory into practice}, 41(4):226--232.

\bibitem[{Mikolov et~al.(2013{\natexlab{a}})Mikolov, Chen, Corrado, and
  Dean}]{DBLP:journals/corr/abs-1301-3781}
Tom{\'{a}}s Mikolov, Kai Chen, Greg Corrado, and Jeffrey Dean.
  2013{\natexlab{a}}.
\newblock Efficient estimation of word representations in vector space.
\newblock In \emph{1st International Conference on Learning Representations,
  {ICLR} 2013, Scottsdale, Arizona, USA, May 2-4, 2013, Workshop Track
  Proceedings}.

\bibitem[{Mikolov et~al.(2013{\natexlab{b}})Mikolov, Le, and
  Sutskever}]{mikolov2013exploiting}
Tomas Mikolov, Quoc~V Le, and Ilya Sutskever. 2013{\natexlab{b}}.
\newblock Exploiting similarities among languages for machine translation.
\newblock \emph{arXiv preprint arXiv:1309.4168}.

\bibitem[{Montague et~al.(1970)}]{montague1970universal}
Richard Montague et~al. 1970.
\newblock Universal grammar.
\newblock \emph{1974}, pages 222--46.

\bibitem[{Nag et~al.(2020)Nag, Kale, Lakshminarasimhan, and
  Singhavi}]{nag2020incorporating}
Sreyashi Nag, Mihir Kale, Varun Lakshminarasimhan, and Swapnil Singhavi. 2020.
\newblock \href {https://openreview.net/forum?id=B1ecYsqSuN} {Incorporating
  bilingual dictionaries for low resource semi-supervised neural machine
  translation}.
\newblock In \emph{International Conference on Learning Representations,
  Learning with Limited Labeled Data}.

\bibitem[{Nguyen and Grieve(2020)}]{nguyen-grieve-2020-word}
Dong Nguyen and Jack Grieve. 2020.
\newblock \href {https://doi.org/10.18653/v1/2020.coling-main.75} {Do word
  embeddings capture spelling variation?}
\newblock In \emph{Proceedings of the 28th International Conference on
  Computational Linguistics}, pages 870--881, Barcelona, Spain (Online).
  International Committee on Computational Linguistics.

\bibitem[{Nye et~al.(2020)Nye, Solar-Lezama, Tenenbaum, and
  Lake}]{NEURIPS2020_7a685d9e}
Maxwell Nye, Armando Solar-Lezama, Josh Tenenbaum, and Brenden~M Lake. 2020.
\newblock \href
  {https://proceedings.neurips.cc/paper/2020/file/7a685d9edd95508471a9d3d6fcace432-Paper.pdf}
  {Learning compositional rules via neural program synthesis}.
\newblock In \emph{Advances in Neural Information Processing Systems},
  volume~33, pages 10832--10842. Curran Associates, Inc.

\bibitem[{Okebukola and Jegede(1988)}]{okebukola1988cognitive}
Peter~Akinsola Okebukola and Olugbemiro~J Jegede. 1988.
\newblock Cognitive preference and learning mode as determinants of meaningful
  learning through concept mapping.
\newblock \emph{Science Education}, 72(4):489--500.

\bibitem[{Oren et~al.(2020)Oren, Herzig, Gupta, Gardner, and
  Berant}]{oren-etal-2020-improving}
Inbar Oren, Jonathan Herzig, Nitish Gupta, Matt Gardner, and Jonathan Berant.
  2020.
\newblock \href {https://doi.org/10.18653/v1/2020.findings-emnlp.225}
  {Improving compositional generalization in semantic parsing}.
\newblock In \emph{Findings of the Association for Computational Linguistics:
  EMNLP 2020}, pages 2482--2495, Online. Association for Computational
  Linguistics.

\bibitem[{Ott et~al.(2019)Ott, Edunov, Baevski, Fan, Gross, Ng, Grangier, and
  Auli}]{ott2019fairseq}
Myle Ott, Sergey Edunov, Alexei Baevski, Angela Fan, Sam Gross, Nathan Ng,
  David Grangier, and Michael Auli. 2019.
\newblock fairseq: A fast, extensible toolkit for sequence modeling.
\newblock In \emph{Proceedings of NAACL-HLT 2019: Demonstrations}.

\bibitem[{Papineni et~al.(2002)Papineni, Roukos, Ward, and
  Zhu}]{papineni2002bleu}
Kishore Papineni, Salim Roukos, Todd Ward, and Wei-Jing Zhu. 2002.
\newblock Bleu: a method for automatic evaluation of machine translation.
\newblock In \emph{Proceedings of the 40th annual meeting of the Association
  for Computational Linguistics}, pages 311--318.

\bibitem[{Pascanu et~al.(2013)Pascanu, Mikolov, and
  Bengio}]{10.5555/3042817.3043083}
Razvan Pascanu, Tomas Mikolov, and Yoshua Bengio. 2013.
\newblock On the difficulty of training recurrent neural networks.
\newblock In \emph{Proceedings of the 30th International Conference on
  International Conference on Machine Learning - Volume 28}, ICML’13, page
  III–1310–III–1318. JMLR.org.

\bibitem[{Patel et~al.(2022)Patel, Bhattamishra, Blunsom, and
  Goyal}]{patel-etal-2022-revisiting}
Arkil Patel, Satwik Bhattamishra, Phil Blunsom, and Navin Goyal. 2022.
\newblock \href {https://doi.org/10.18653/v1/2022.acl-short.46} {Revisiting the
  compositional generalization abilities of neural sequence models}.
\newblock In \emph{Proceedings of the 60th Annual Meeting of the Association
  for Computational Linguistics (Volume 2: Short Papers)}, pages 424--434,
  Dublin, Ireland. Association for Computational Linguistics.

\bibitem[{Post(2018)}]{post-2018-call}
Matt Post. 2018.
\newblock \href {https://doi.org/10.18653/v1/W18-6319} {A call for clarity in
  reporting {BLEU} scores}.
\newblock In \emph{Proceedings of the Third Conference on Machine Translation:
  Research Papers}, pages 186--191, Brussels, Belgium. Association for
  Computational Linguistics.

\bibitem[{Prechelt(1998)}]{prechelt1998early}
Lutz Prechelt. 1998.
\newblock Early stopping-but when?
\newblock In \emph{Neural Networks: Tricks of the trade}, pages 55--69.
  Springer.

\bibitem[{Rubenstein and Goodenough(1965)}]{10.1145/365628.365657}
Herbert Rubenstein and John~B. Goodenough. 1965.
\newblock \href {https://doi.org/10.1145/365628.365657} {Contextual correlates
  of synonymy}.
\newblock \emph{Commun. ACM}, 8(10):627–633.

\bibitem[{Schuster and Paliwal(1997)}]{schuster1997bidirectional}
Mike Schuster and Kuldip~K Paliwal. 1997.
\newblock Bidirectional recurrent neural networks.
\newblock \emph{IEEE transactions on Signal Processing}, 45(11):2673--2681.

\bibitem[{Shaffer(1989)}]{shaffer1989comparison}
Constance Shaffer. 1989.
\newblock A comparison of inductive and deductive approaches to teaching
  foreign languages.
\newblock \emph{The Modern Language Journal}, 73(4):395--403.

\bibitem[{Shaw et~al.(2021)Shaw, Chang, Pasupat, and
  Toutanova}]{shaw-etal-2021-compositional}
Peter Shaw, Ming-Wei Chang, Panupong Pasupat, and Kristina Toutanova. 2021.
\newblock \href {https://doi.org/10.18653/v1/2021.acl-long.75} {Compositional
  generalization and natural language variation: Can a semantic parsing
  approach handle both?}
\newblock In \emph{Proceedings of the 59th Annual Meeting of the Association
  for Computational Linguistics and the 11th International Joint Conference on
  Natural Language Processing (Volume 1: Long Papers)}, pages 922--938, Online.
  Association for Computational Linguistics.

\bibitem[{Sinha et~al.(2019)Sinha, Sodhani, Dong, Pineau, and
  Hamilton}]{sinha-etal-2019-clutrr}
Koustuv Sinha, Shagun Sodhani, Jin Dong, Joelle Pineau, and William~L.
  Hamilton. 2019.
\newblock \href {https://doi.org/10.18653/v1/D19-1458} {{CLUTRR}: A diagnostic
  benchmark for inductive reasoning from text}.
\newblock In \emph{Proceedings of the 2019 Conference on Empirical Methods in
  Natural Language Processing and the 9th International Joint Conference on
  Natural Language Processing (EMNLP-IJCNLP)}, pages 4506--4515, Hong Kong,
  China. Association for Computational Linguistics.

\bibitem[{Srinivasan et~al.(2017)Srinivasan, Ioannis, Alvin, Jayant, and
  Luke}]{data-atis-geography-scholar}
Iyer Srinivasan, Konstas Ioannis, Cheung Alvin, Krishnamurthy Jayant, and
  Zettlemoyer Luke. 2017.
\newblock Learning a neural semantic parser from user feedback.
\newblock In \emph{Proceedings of the 55th Annual Meeting of the Association
  for Computational Linguistics (Volume 1: Long Papers)}, pages 963--973.

\bibitem[{Srivastava et~al.(2014)Srivastava, Hinton, Krizhevsky, Sutskever, and
  Salakhutdinov}]{srivastava2014dropout}
Nitish Srivastava, Geoffrey Hinton, Alex Krizhevsky, Ilya Sutskever, and Ruslan
  Salakhutdinov. 2014.
\newblock Dropout: a simple way to prevent neural networks from overfitting.
\newblock \emph{The journal of machine learning research}, 15(1):1929--1958.

\bibitem[{Stanojevi{\'c} et~al.(2009)}]{stanojevic2009cognitive}
Maja Stanojevi{\'c} et~al. 2009.
\newblock Cognitive synonymy: A general overview.
\newblock \emph{FACTA UNIVERSITATIS-Linguistics and Literature}, 7(2):193--200.

\bibitem[{Sutskever et~al.(2014)Sutskever, Vinyals, and
  Le}]{sutskever2014sequence}
Ilya Sutskever, Oriol Vinyals, and Quoc~V Le. 2014.
\newblock Sequence to sequence learning with neural networks.
\newblock In Z.~Ghahramani, M.~Welling, C.~Cortes, N.~D. Lawrence, and K.~Q.
  Weinberger, editors, \emph{Advances in Neural Information Processing Systems
  27}, pages 3104--3112. Curran Associates, Inc.

\bibitem[{Thornbury(1999)}]{thornbury1999teach}
Scott Thornbury. 1999.
\newblock \emph{How to teach grammar}, volume~3.
\newblock Longman Harlow.

\bibitem[{Vaswani et~al.(2017)Vaswani, Shazeer, Parmar, Uszkoreit, Jones,
  Gomez, Kaiser, and Polosukhin}]{NIPS2017_7181}
Ashish Vaswani, Noam Shazeer, Niki Parmar, Jakob Uszkoreit, Llion Jones,
  Aidan~N Gomez, \L~ukasz Kaiser, and Illia Polosukhin. 2017.
\newblock Attention is all you need.
\newblock In I.~Guyon, U.~V. Luxburg, S.~Bengio, H.~Wallach, R.~Fergus,
  S.~Vishwanathan, and R.~Garnett, editors, \emph{Advances in Neural
  Information Processing Systems 30}, pages 5998--6008. Curran Associates, Inc.

\bibitem[{Wang and He(2020)}]{wang2020birre}
Chengyu Wang and Xiaofeng He. 2020.
\newblock Birre: learning bidirectional residual relation embeddings for
  supervised hypernymy detection.
\newblock In \emph{Proceedings of the 58th Annual Meeting of the Association
  for Computational Linguistics}, pages 3630--3640.

\bibitem[{Williams and Zipser(1989)}]{williams1989learning}
Ronald~J Williams and David Zipser. 1989.
\newblock A learning algorithm for continually running fully recurrent neural
  networks.
\newblock \emph{Neural computation}, 1(2):270--280.

\bibitem[{Wu et~al.(2019)Wu, Fan, Baevski, Dauphin, and Auli}]{wu2018pay}
Felix Wu, Angela Fan, Alexei Baevski, Yann Dauphin, and Michael Auli. 2019.
\newblock Pay less attention with lightweight and dynamic convolutions.
\newblock In \emph{International Conference on Learning Representations}.

\bibitem[{Xie et~al.(2020)Xie, Dai, Hovy, Luong, and Le}]{xie2019unsupervised}
Qizhe Xie, Zihang Dai, Eduard~H. Hovy, Thang Luong, and Quoc Le. 2020.
\newblock \href
  {https://proceedings.neurips.cc/paper/2020/hash/44feb0096faa8326192570788b38c1d1-Abstract.html}
  {Unsupervised data augmentation for consistency training}.
\newblock In \emph{Advances in Neural Information Processing Systems}.

\bibitem[{Yin et~al.(2021)Yin, Fang, Neubig, Pauls, Platanios, Su, Thomson, and
  Andreas}]{yin-etal-2021-compositional}
Pengcheng Yin, Hao Fang, Graham Neubig, Adam Pauls, Emmanouil~Antonios
  Platanios, Yu~Su, Sam Thomson, and Jacob Andreas. 2021.
\newblock \href {https://doi.org/10.18653/v1/2021.naacl-main.225}
  {Compositional generalization for neural semantic parsing via span-level
  supervised attention}.
\newblock In \emph{Proceedings of the 2021 Conference of the North American
  Chapter of the Association for Computational Linguistics: Human Language
  Technologies}, pages 2810--2823, Online. Association for Computational
  Linguistics.

\bibitem[{Yu et~al.(2020{\natexlab{a}})Yu, Shen, Ma, Jia, Chen, and
  Lu}]{yu2020synet}
Jiale Yu, Yongliang Shen, Xinyin Ma, Chenghao Jia, Chen Chen, and Weiming Lu.
  2020{\natexlab{a}}.
\newblock Synet: Synonym expansion using transitivity.
\newblock In \emph{Proceedings of the 2020 Conference on Empirical Methods in
  Natural Language Processing: Findings}, pages 1961--1970.

\bibitem[{Yu et~al.(2020{\natexlab{b}})Yu, Li, Shen, Feng, Sun, and
  Zhang}]{yu2020steam}
Yue Yu, Yinghao Li, Jiaming Shen, Hao Feng, Jimeng Sun, and Chao Zhang.
  2020{\natexlab{b}}.
\newblock Steam: Self-supervised taxonomy expansion with mini-paths.
\newblock In \emph{Proceedings of the 26th ACM SIGKDD International Conference
  on Knowledge Discovery \& Data Mining}, pages 1026--1035.

\bibitem[{Zelle and Mooney(1996)}]{data-geography-original}
John~M. Zelle and Raymond~J. Mooney. 1996.
\newblock Learning to parse database queries using inductive logic programming.
\newblock In \emph{Proceedings of the Thirteenth National Conference on
  Artificial Intelligence - Volume 2}, pages 1050--1055.

\end{thebibliography}
